%% file: main.tex
\documentclass{article}

\usepackage{iclr2027_conference,times}
\input{math_commands.tex}

\usepackage{amsmath}
\usepackage{booktabs}
\usepackage{multirow}
\usepackage{enumitem}
\usepackage{float}
\usepackage{graphicx}
\usepackage{hyperref}

\usepackage{microtype}
\usepackage{url}
\usepackage{xcolor}
\usepackage{fvextra}
\usepackage{needspace}
\usepackage{tcolorbox}
\tcbuselibrary{breakable,skins}
\DefineVerbatimEnvironment{ConstructionPrompt}{Verbatim}{%
  fontsize=\fontsize{8}{9.5}\selectfont,baselinestretch=1,
  breaklines=true,breakanywhere=true,
  breaksymbolleft={},breaksymbolright={},
  breakanywheresymbolpre={},breakanywheresymbolpost={},
  breakautoindent=true,breakindent=0pt,
  frame=none,tabsize=2,listparameters={\setlength{\topsep}{0pt}}
}
\newcommand{\PromptTitle}[1]{\def\CurrentPromptTitle{#1}}
\PromptTitle{Prompt}
\tcolorboxenvironment{ConstructionPrompt}{%
  enhanced,breakable,colback=black!2,colframe=black!30,
  title=\CurrentPromptTitle,
  title after break={\CurrentPromptTitle\space (continued)},
  colbacktitle=black!8,coltitle=black,fonttitle=\small\bfseries,
  titlerule=0pt,toptitle=1mm,bottomtitle=1mm,
  boxrule=0.35pt,arc=1mm,boxsep=0pt,
  left=2mm,right=2mm,top=1.5mm,bottom=1.5mm,
  before skip=5pt,after skip=7pt
}

\setlist[itemize]{leftmargin=*, topsep=3pt, partopsep=0pt, parsep=0pt, itemsep=3pt}

\title{Beyond Perfect Teammates: Multi-Agent \\ Collaboration with Unreliable Colleagues}

\usepackage{latexml}
\iflatexml
\begingroup
\usepackage{amsaddr}
\usepackage{authblk}
\author[chen]{Chen Huang\thanks{Equal contribution.}}
\affil[chen]{Singapore University of Technology and Design}
\email{chen\_huang@mymail.sutd.edu.sg}
\author[yiyang]{Yiyang Zhang\thanks{Equal contribution.}}
\affil[yiyang]{Singapore University of Technology and Design}
\email{yiyang\_zhang@mymail.sutd.edu.sg}
\author[yuhao]{Yuhao Wu}
\affil[yuhao]{Singapore University of Technology and Design}
\author[yang]{Yang Deng}
\affil[yang]{Singapore Management University}
\author[wenxuan]{Wenxuan Zhang\thanks{Corresponding author.}}
\affil[wenxuan]{Singapore University of Technology and Design}
\email{wxzhang@sutd.edu.sg}
\global\let\emailaddrname\emailaddrname
\endgroup
\else
\author{\small Chen Huang$^{1,}$\thanks{Equal contribution.} \quad Yiyang Zhang$^{1,}$\footnotemark[1] \quad Yuhao Wu$^1$ \quad Yang Deng$^2$ \quad Wenxuan Zhang$^{1,}$\thanks{Corresponding author.}\\
{\normalfont\small $^1$Singapore University of Technology and Design}\\
{\normalfont\small $^2$Singapore Management University}\\
{\normalfont\small\texttt{\{chen\_huang,yiyang\_zhang\}@mymail.sutd.edu.sg, wxzhang@sutd.edu.sg}}}
\fi

\iclrfinalcopy

\begin{document}

\maketitle
\lhead{\footnotesize Beyond Perfect Teammates: Multi-Agent Collaboration with Unreliable Colleagues}

\input{sections/00_Abstract}
\input{sections/01_Introduction}
\input{sections/02_Related_Work}
\input{sections/04_Task_Construction}
\input{sections/05_Experimental_Setup}
\input{sections/06_Results_and_Analysis}

\input{sections/08_Conclusion}
\bibliography{iclr2027_conference}
\bibliographystyle{iclr2027_conference}

\appendix

\input{sections/09_Appendix}
\input{sections/12_Causal_Intervention}
\input{sections/13_Multiple_Unreliable_Colleagues}

\end{document}

%% file: math_commands.tex
\usepackage{amsmath,amsfonts,bm}

\def\eqref#1{equation~\ref{#1}}

\def\1{\bm{1}}

\DeclareMathAlphabet{\mathsfit}{\encodingdefault}{\sfdefault}{m}{sl}
\SetMathAlphabet{\mathsfit}{bold}{\encodingdefault}{\sfdefault}{bx}{n}



%% file: sections/00_Abstract.tex
\begin{abstract}
LLM-based multi-agent collaboration is increasingly used to solve complex tasks by dividing work among specialized agents and integrating complementary expertise. Yet existing systems and evaluations largely rest on an implicit perfect-colleague assumption: collaborators provide correct and complete information and respond promptly to requests. In realistic teamwork, however, collaborators may return information that does not answer the intended question, provide insufficient detail to support a decision, or fail to respond to a request. To separate task-solving competence from the ability to recover an unreliable evidence-gathering process, we introduce the Unreliable Colleagues Benchmark (\textsc{UnCoBench}), which organizes behavioral failures through \emph{Misaligned}, \emph{Incomplete}, and \emph{Missed} settings. The main agent need to verify contextual correctness, diagnose information gap and track unresolved dependencies in order to obtain the correct evidence for solving the task. We also introduce a verifiable and scalable automated generation pipeline that pairs LLM-based scenario generation with executable checks of answer uniqueness and role necessity. Results across thirteen models show that similar reliable-setting performance can result in different robustness to unreliable evidence. Trace analysis reveals frequent omissions of corrective follow-up, and the recovery behavior involves inefficient waiting and redundant queries. These findings show that robust collaboration requires not only sound reasoning, but also targeted action to diagnose evidence failures, recover relevant information, and integrate it into decisions.
Our data and source code are available at \url{https://github.com/iNLP-Lab/UnCoBench}.
\end{abstract}

%% file: sections/01_Introduction.tex
\section{Introduction}
\label{sec:introduction}

\begin{figure}[t]
  \centering
  \setlength{\abovecaptionskip}{2pt}  
  \includegraphics[width=\linewidth]{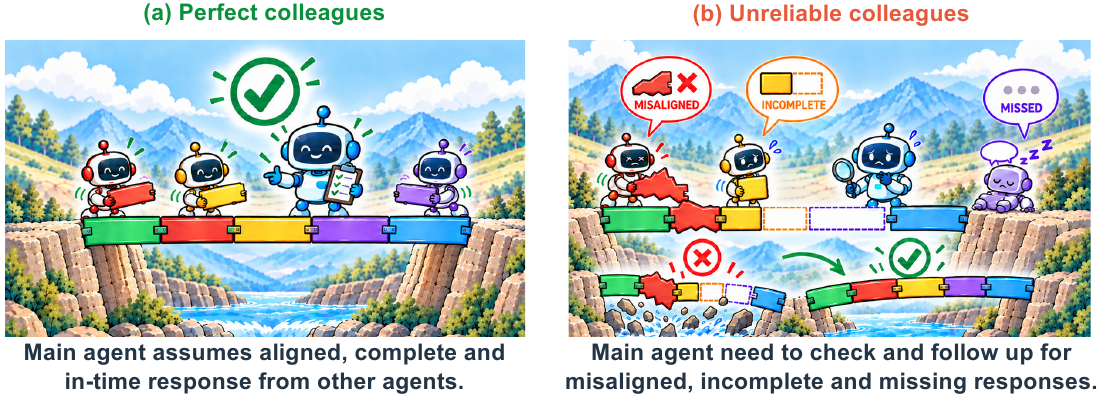}
  \caption{Behavior difference between perfect and unreliable colleagues in multi-agent collaboration.}
  \label{fig:behavior-difference}
  \vspace{-2em}
\end{figure}

LLM-based multi-agent systems increasingly tackle complex tasks by distributing work across specialized agents, exchanging intermediate results, and integrating complementary expertise~\citep{yun2026graph,li2025agent,chen2025internet}. Recent systems operationalize this idea through explicit task decomposition, programmable communication, and structured workflows~\citep{li2023camel,chen2024agentverse,wu2024autogen,hong2024metagpt,qian2024chatdev}, while benchmarks examine communication topology, coordination, and distributed private evidence~\citep{zhu2025multiagentbench,liu2024iagents,li2026hiddenbench}. Yet both design and evaluation largely rest on an implicit perfect-colleague assumption as in Figure~\ref{fig:behavior-difference}(a): the responses of specialized role agents participating in the collaboration are expected to be aligned with target, complete in data scope, and timely in response. 

Realistic human collaboration, however, violates this assumption in various ways~\citep{reason2000human,lingard2004communication}. A colleague may provide a plausible figure from the wrong context, omit a detail needed to check a threshold, or miss the request because they are distracted or fail to notice it~\citep{lingard2004communication}. None of these behaviors is deceptive, yet each shifts the burden onto the requesting agent: as shown in Figure~\ref{fig:behavior-difference}(b), a coordinator must verify that the evidence it receives matches the scope it asked about, recognize when a decision-critical detail is absent, and schedule follow-ups rather than wait indefinitely or repeatedly query an unavailable colleague. 
However, it remains unclear how well current LLM agents can perform these behaviors, because existing evaluations largely measure whether agents can solve tasks when collaborators respond perfectly~\citep{cemri2025masfail}, rather than whether they can recover when collaboration itself becomes unreliable. As a result, strong task performance under ideal multi-agent settings may obscure important weaknesses in verification, clarification, and recovery.
This leads to a realistic question: \textbf{How well can an agent complete a collaborative task when its colleagues exhibit unreliable behavior?}

To answer this question, we introduce the Unreliable Colleagues Benchmark (\textsc{UnCoBench}), a controlled benchmark designed to separate underlying task-solving competence from the ability to recover an unreliable evidence-gathering process. 
We organize the behavioral failures around three fundamental properties of the evidence-delivery process—contextual alignment, completeness, and responsiveness—and introduce three unreliable settings. Under \textit{Misaligned}, a collaborator may return plausible evidence for the wrong context, requiring the main agent to verify its applicability and seek a correction; \textit{Incomplete} provide correct but insufficient evidence, requiring the agent to diagnose the information gap and request clarification; \textit{Missed} fails to answer an initial request, requiring the agent to track unresolved dependencies, schedule follow-up, and recontact the relevant role.


We formalize these three forms of unreliability in \textsc{UnCoBench} as a benchmark of 1,000 paired, structured decision-making tasks through a scalable, automated, and verifiable pipeline.  Instead of  generating complete tasks end-to-end~\citep{zhu2025multiagentbench,liu2025mcpeval,wang2025odysseybench,PARTNR}, an LLM first generates diverse structured scenarios and role specifications, after which programs instantiate the underlying evidence and verify answer uniqueness and role necessity. The response policies of role agents are then systematically adapted to each unreliable setting, introducing the designated misaligned, incomplete, or missed behavior while keeping the underlying task and ground truth fixed. Finally, an independent LLM validates the rendered natural-language instances to ensure that the tasks remain solvable and the verified gold answer can be consistently recovered.

We evaluate \textsc{UnCoBench} across thirteen open- and closed-source models, examining both final task outcomes and the interaction pathways through which models respond to unreliable collaborator evidence. First, models with similar performance under reliable collaboration can exhibit significant different evidence-recovery behavior. GPT-5.6 Sol and Claude Opus 5 achieve similarly high accuracy under \textit{Reliable}, at 98.7\% and 96.7\%, respectively, but their performance diverges sharply under \textit{Misaligned}, falling to 45.8\% and 90.7\%.
Second, trace-level analysis suggests that a key bottleneck arises before evidence can be successfully recovered, rather than solely at the stage of final answer integration. Under \textit{Misaligned}, more than half of GPT-5.6 Sol episodes contain no corrective follow-up, revealing limited ability to actively verify questionable evidence and initiate targeted recovery.
Moreover, additional interaction is not uniformly productive: recovery from missed responses can involve both successful recontact and inefficient waiting or redundant queries. These results show that robust multi-agent collaboration requires more than general reasoning ability: agents must detect when provided evidence is misaligned, incomplete, or missing, diagnose the underlying failure, and take targeted actions to recover and correctly integrate the evidence needed for the final decision.

To summarize, our work makes the following key contributions:
\begin{itemize}
    \item We formalize a controlled diagnostic setting for multi-agent collaboration with unreliable colleagues, using \textit{Misaligned}, \textit{Incomplete}, and \textit{Missed} profiles to vary the contextual alignment, completeness, and responsiveness of evidence, separating task competence from evidence-recovery capability.
    \item We instantiate this setting in \textsc{UnCoBench}, a benchmark of 1,000 paired tasks constructed with executable checks of answer uniqueness, role necessity, perturbation validity, and solvability.
    \item Across models, we show that similar \textit{Reliable} performance does not necessarily translate into similar recovery behavior. Trace analysis identifies absent or non-targeted corrective follow-up and inefficient recovery from non-response as prominent observable bottlenecks.
\end{itemize}

%% file: sections/02_Related_Work.tex
\section{Related Work}
\label{sec:related-work}

\paragraph{Collaboration under distributed information.} Evaluations of multi-agent collaboration examine both task completion and the interactions that support it, including coordination protocols and intermediate milestones \citep{zhu2025multiagentbench}. Under information asymmetry, effective interaction also requires understanding what other participants know and eliciting facts that are not initially shared. Research on knowledge attribution and proactive information exchange studies complementary aspects of this process \citep{kim2023fantom,liu2024iagents}, while HiddenBench compares information pooling in hidden-profile tasks with full-information controls \citep{li2026hiddenbench}. These studies motivate distinguishing the availability of knowledge within a team from its use in a decision. \textsc{UnCoBench} examines this distinction under unreliable collaboration and explicitly varies collaborator response reliability, testing whether a coordinator obtains the required evidence when colleagues return mis-scoped, incomplete information, or fail to respond.

\paragraph{Controlled diagnosis of multi-agent failures.} Multi-agent failure analysis connects recurring error patterns to their consequences and the possibility of recovery. MAST categorizes specification, coordination, and verification failures, including omitted information and failures to seek clarification \citep{cemri2025masfail}. Attribution methods sharpen this diagnosis by examining which errors matter for the outcome: AgenTracer uses counterfactual replay and programmed fault injection to identify agents and steps responsible for failed trajectories \citep{zhang2026agentracer}. Complementary resilience evaluations use controlled faults to test how systems respond and recover, perturbing collaborator messages as well as prompts and message routing \citep{huang2025resilience,jia2026masfire}. \textsc{UnCoBench} makes collaboration necessary by distributing the facts needed to determine a unique answer across role agents' private records, and evaluates whether the main agent can detect problematic replies, recover those facts through targeted follow-up, and use them to reach the correct decision.

\paragraph{Evidence adequacy and interactive recovery.} A response can be plausible without being reliable or sufficient, as studies of silent tool errors and retrieval robustness demonstrate \citep{sun2024ToolsFail,chen2024RGB}. Recovery therefore connects evidence assessment to the choice of a subsequent action. Tool-recovery tasks and evaluations with user- and tool-side noise examine this connection under perturbations that preserve task solvability \citep{wang2025HellOrHighWater,wang2026AgentNoiseBench}, while clarification research studies how dialogue resolves underspecified requests and false premises \citep{zhao2026AskBench}. \textsc{UnCoBench} retains a route to the required evidence through further interaction, using recovery traces alongside final accuracy to help distinguish unsuccessful information gathering from errors after the evidence has been recovered.

%% file: sections/04_Task_Construction.tex
\section{Automated Benchmark Construction}
\label{sec:construction}


In \textsc{UnCoBench}, each task involves a main agent and several specialized role agents, with each role agent holding private records. The main agent is the subject of evaluation and must gather and integrate evidence from every necessary role to solve a decision problem defined by three candidates, a set of eligibility conditions, and a numerical calculation rule. The main agent must determine whether any candidate satisfies the eligibility conditions and, if so, compute the corresponding numerical result. To evaluate how colleague reliability affects the main agent's performance, each task has four settings: \textbf{Reliable}, \textbf{Misaligned}, \textbf{Incomplete}, and \textbf{Missed}.

\begin{figure}[tp]
  \centering
  \includegraphics[width=\linewidth]{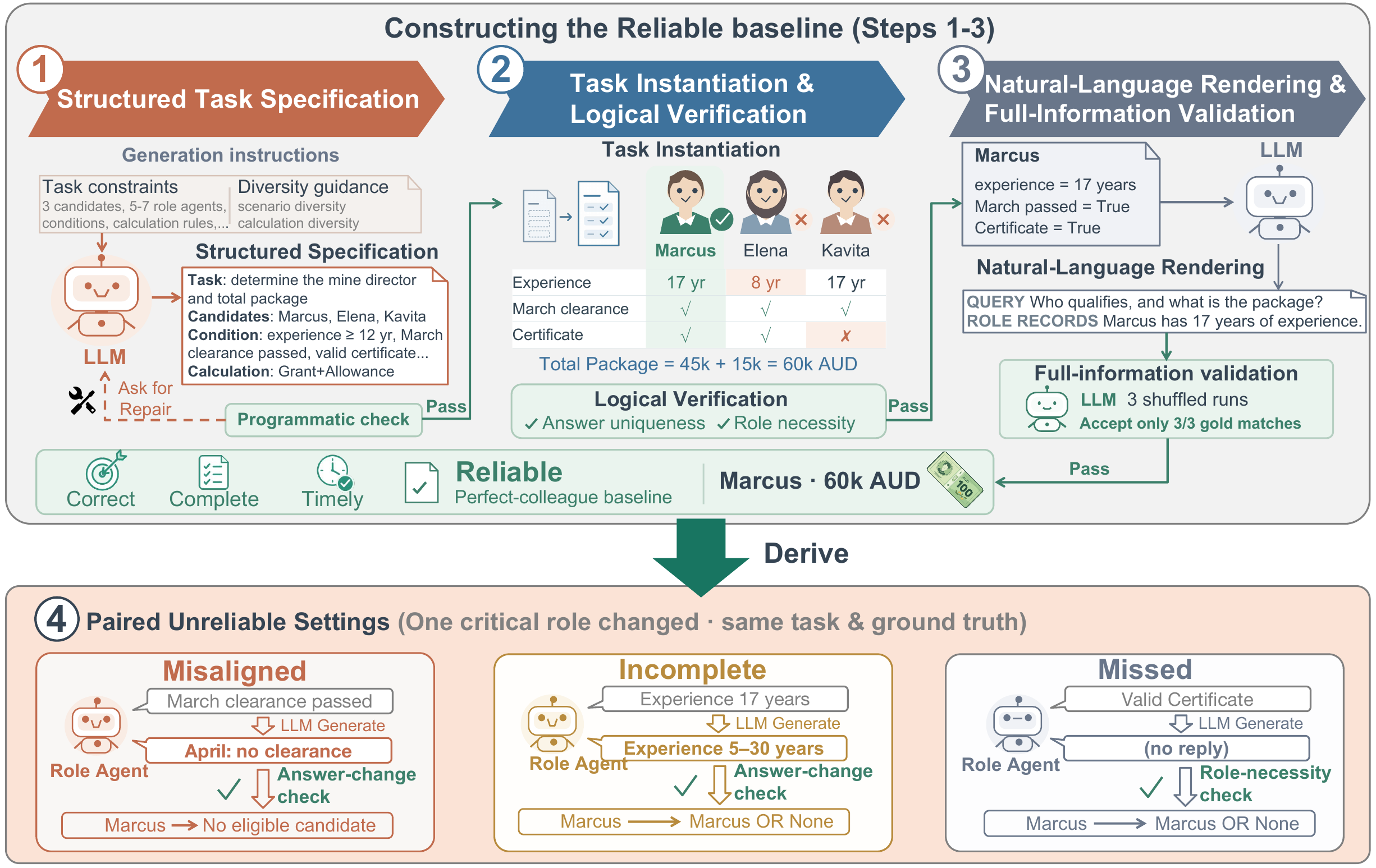}
  \caption{Construction pipeline for \textsc{UnCoBench}. The first three steps produce structured specifications, logically verified instances, and validated natural-language tasks, establishing the Reliable baseline. The final step then derives three paired unreliable settings.}
  \label{fig:construction-pipeline}
\end{figure}

As shown in Figure~\ref{fig:construction-pipeline}, we construct these tasks through a modular automated pipeline that combines LLM-based generation with programmatic verification. First, an LLM generates diverse structured task specifications. Second, a program instantiates and logically verifies these specifications. Third, LLM-based natural-language rendering and full-information validation establish the Reliable perfect-colleague baseline. Finally, the pipeline derives the paired unreliable settings by changing the response policy of one necessary role. The following subsections describe each step in detail. 

\subsection{Structured Task Specification}
\label{sec:task-specification}

In the first step, an LLM generates a structured task specification following task constraints and diversity guidance. The constraints define the overall task structure, including the candidates, agent roles, eligibility conditions, and numerical calculation rule, while diversity guidance encourages variation in both scenarios and numerical computation. To encourage scenario diversity, the LLM receives eight reference scenarios and summaries of recently accepted tasks. It can adapt these references or create new scenarios while avoiding repetition. To diversify computation, the program selects one of eight supported operations, favoring less frequently used ones, and the LLM assigns it a scenario-appropriate interpretation. More details are provided in Appendix~\ref{app:Specification}.

Each specification is checked programmatically for condition distribution across roles, operand ownership, value domains, and units. Invalid specifications are returned to the LLM with validation feedback for repair. Only those that pass proceed to task instantiation.

\subsection{Task Instantiation and Logical Verification}
\label{sec:task-instantiation}

The second step converts each specification into a concrete task instance. A program assigns values from finite domains to candidate-specific evidence held by the role agents, ensures that at most one candidate qualifies, and computes the ground-truth answer. In the example illustrated in Figure~\ref{fig:construction-pipeline}, only Marcus satisfies all eligibility conditions, with a total package of $45{,}000 + 15{,}000 = 60{,}000$~AUD. The grant and allowance values are also part of the evidence but are omitted from the figure.

After instantiation, the program verifies two logical properties. Answer uniqueness requires the complete evidence to determine a unique answer. Role necessity requires that removing the evidence held by any necessary role leaves the answer undetermined. Together, these checks ensure that the task is solvable with complete evidence and that every necessary role contributes indispensable information. The verification criteria are detailed in Appendix~\ref{app:logical-verification}.

\subsection{Natural-Language Rendering and Full-Information Validation}
\label{sec:task-rendering}

In the third step, an LLM renders each logically verified instance as a natural-language task, including the task background and query for the main agent and private records for each role agent. The rendering preserves the assigned facts, eligibility conditions, and evidence contexts to maintain the underlying decision problem.

We then perform full-information validation to assess whether the rendered task is solvable with complete evidence. A separate validation LLM receives the query, eligibility conditions, and all role records, and solves the task in three independent runs, each with a shuffled evidence order. A task is retained for the Reliable setting only if all three answers match the ground truth. This empirical check of natural-language solvability complements the logical verification of structured instances, thus supporting the reliability of the automated construction pipeline. Rendering and validation prompts are provided in Appendix~\ref{app:rendering-validation}.


\subsection{Unreliable Settings and Recovery Mechanisms}
\label{sec:unreliable-construction}

The preceding steps establish Reliable tasks as the perfect-colleague baseline, and a concrete example is provided in Appendix~\ref{app:example-setup}. Then the fourth step derives the paired \textit{Misaligned}, \textit{Incomplete}, and \textit{Missed} settings by modifying exactly one necessary role's response policy while preserving the underlying decision problem and ground-truth answer. These three unreliable settings capture common failure patterns in real multi-agent collaboration \cite{zhang2025agent}.

\textbf{Misaligned} introduces an initial reply with a mismatched context. An LLM specifies a mismatch in location, time period, or measurement definition, and a program assigns alternative values and verifies that using them changes the inferred answer. For example, reporting a failed April clearance instead of the required March clearance changes the inferred decision from Marcus to no eligible candidate. The main agent must detect the mismatch and explicitly request evidence for the correct context. More details can be found in Appendix~\ref{app:misaligned-template}.

\textbf{Incomplete} constructs evidence that remains correct but is insufficient to determine the answer. An LLM generates an initial reply that omits a numerical value or provides only a coarse bound or interval. A program verifies that the reply remains truthful but no longer uniquely determines the answer. For example, reporting ``5--30 years'' instead of 17 years leaves eligibility under a 12-year experience requirement undetermined. The main agent must identify the information gap and request the precise value. Appendix~\ref{app:incomplete-template} provides a more detailed example. 

\textbf{Missed} models a collaborator who overlooks the initial request leaving the underlying evidence unchanged but providing no response. The role-necessity check in Step~2 ensures that the remaining evidence is insufficient to derive the original ground-truth answer. The main agent must therefore detect the missing response and follow up to obtain the required evidence.

Through this pipeline, we construct \textsc{UnCoBench} with 1,000 task groups, each comprising \textit{Reliable}, \textit{Misaligned}, \textit{Incomplete}, and \textit{Missed} variants that share the same underlying decision problem and ground-truth answer.

\subsection{Generation Diversity and Validity}
We further analyze the diversity and validity of \textsc{UnCoBench} to assess whether the proposed construction pipeline produces broad yet well-formed tasks rather than merely scaling up a narrow generation pattern. In terms of diversity, the retained tasks span 14 application domains and eight numerical operations, with relatively balanced coverage across both dimensions, suggesting that the pipeline does not collapse toward a small set of topics or reasoning templates. In terms of validity, the automated verification stages reject 296 of 1,296 candidate task groups (22.8\%), retaining only instances that satisfy structural, answer-uniqueness, role-necessity, and full-information solvability checks; all retained tasks also recover the programmatically computed gold answer across repeated shuffled validation runs. We additionally conduct human evaluation of logical coherence and task solvability, where strong agreement with human judgments further supports the effectiveness of the construction pipeline. Detailed analyses are provided in Appendix~\ref{app:benchmark-validity}.

%% file: sections/05_Experimental_Setup.tex
\section{Experimental Setup}
\label{sec:experiments}

\input{tables/main_results}

\subsection{Models and protocol}

We evaluate 13 main-agent models on \textsc{UnCoBench}, covering both closed-source and open-weight groups. The closed-source set comprises GPT-5.6 Sol~\citep{openai2026gpt56sol}, Claude Opus 5~\citep{anthropic2026claudeopus5}, Gemini 3.8 Flash~\citep{google2026gemini38flash}, and Qwen4-Max~\citep{qwen2026qwen4max}. The open-weight set comprises Hy4~\citep{tencent2026hy4}, Hy4-Flash~\citep{tencent2026hy4flash}, DeepSeek-V4.1-Flash~\citep{deepseek2026v41flash}, GLM-5.3-Flash~\citep{glm5team2026glm5}, Muse-Glimmer-30B~\citep{meta2026museglimmer}, Gemma 4 31B~\citep{gemmateam2026gemma4}, Mistral Small 4-119A6B~\citep{mistralai2026mistralsmall4}, Qwen3.8-27B~\citep{qwen2026qwen38}, and Qwen3.5-9B~\citep{qwen2026qwen35}.
The closed-source models, Hy4, Hy4-Flash, and DeepSeek-V4.1-Flash are accessed through provider APIs. Locally deployed open-weight models are served using vLLM~\citep{kwon2023vllm}.
In our setting, role agents primarily respond to the main agent according to their assigned profiles, placing relatively limited demands on reasoning capability. Based on a comparison of factual fidelity and information coverage across candidate models, we select DeepSeek-V4-Flash-0731~\citep{deepseek2026v4flash0731} as the role-agent model for its strong instruction-following performance and consistent adherence to the specified response policies. Detailed analysis and comparison of alternative role-agent models are provided in Appendix~\ref{app:role-agent-validity}.

Every main agent receives the same tool schema, role descriptions, and final-answer format.
Specifically, the tool schema allows the main agent to query one or more role agents within a single round and optionally choose to wait without issuing a new request, with each returned response attributed to its source role.
For models accessed through APIs, the maximum output length is left at the provider default and is set to 8,192 tokens for locally deployed models. All models use their respective default reasoning settings and are required to return JSON objects in their response. We evaluate the generated 1,000 groups of paired tasks, each comprising \textit{Reliable}, \textit{Misaligned}, \textit{Incomplete}, and \textit{Missed}, yielding 4,000 tasks per main agent.

\subsection{Evaluation Metrics}
We evaluate performance using task accuracy (Acc.) and interaction rounds (Rounds). A task is counted as correct only when both the final decision and the numerical result match the ground-truth answer. Each round corresponds to one main-agent action and may include queries to multiple role agents, we calculate the average rounds across all tasks.



%% file: tables/main_results.tex
\newcommand{\accdelta}[1]{\hspace{0.15em}\makebox[3.0em][r]{\textcolor{red!65!black}{\scriptsize(#1)}}}
\newcommand{\rounddelta}[1]{\hspace{0.15em}\makebox[3.2em][r]{\textcolor{blue!65!black}{\scriptsize(#1)}}}

\begin{table}[t]
\centering
\caption{Main results on \textsc{UnCoBench}. $\dagger$ and $\ddagger$ denote closed-source and open-weight models respectively. Colored deltas indicate changes from the \textit{Reliable} setting: red for accuracy, measured in percentage points, and blue for mean interaction rounds.}
\label{tab:main-results}
\vspace{0.6em}
\small
\setlength{\tabcolsep}{3pt}
\resizebox{\textwidth}{!}{%
\begin{tabular}{lccrrrrrr}
\toprule
& \multicolumn{2}{c}{Reliable}
& \multicolumn{2}{c}{Misaligned}
& \multicolumn{2}{c}{Incomplete}
& \multicolumn{2}{c}{Missed} \\
\cmidrule(lr){2-3}\cmidrule(lr){4-5}\cmidrule(lr){6-7}\cmidrule(lr){8-9}
Main agent
& \multicolumn{1}{c}{Acc. (\%) $\uparrow$} & \multicolumn{1}{c}{Rounds $\downarrow$}
& \multicolumn{1}{c}{Acc. (\%) $\uparrow$} & \multicolumn{1}{c}{Rounds $\downarrow$}
& \multicolumn{1}{c}{Acc. (\%) $\uparrow$} & \multicolumn{1}{c}{Rounds $\downarrow$}
& \multicolumn{1}{c}{Acc. (\%) $\uparrow$} & \multicolumn{1}{c}{Rounds $\downarrow$} \\
\midrule
GPT-5.6 Sol\textsuperscript{$\dagger$} & 98.7 & 2.18 & 45.8\accdelta{-52.9} & 3.06\rounddelta{+0.89} & 77.0\accdelta{-21.7} & 2.93\rounddelta{+0.76} & 97.4\accdelta{-1.3} & 4.97\rounddelta{+2.80} \\
Claude Opus 5\textsuperscript{$\dagger$} & 96.7 & 2.18 & 90.7\accdelta{-6.0} & 3.99\rounddelta{+1.81} & 95.7\accdelta{-1.0} & 3.11\rounddelta{+0.94} & 95.8\accdelta{-0.9} & 3.24\rounddelta{+1.06} \\
Gemini 3.8 Flash\textsuperscript{$\dagger$} & 99.3 & 2.09 & 74.1\accdelta{-25.2} & 3.58\rounddelta{+1.49} & 88.5\accdelta{-10.8} & 2.91\rounddelta{+0.82} & 99.1\accdelta{-0.2} & 3.09\rounddelta{+1.00} \\
Qwen4-Max\textsuperscript{$\dagger$} & 98.2 & 2.41 & 85.1\accdelta{-13.1} & 3.96\rounddelta{+1.55} & 99.2\accdelta{+1.0} & 3.15\rounddelta{+0.74} & 97.7\accdelta{-0.5} & 3.89\rounddelta{+1.48} \\
\midrule
Hy4\textsuperscript{$\ddagger$} & 98.8 & 2.28 & 91.2\accdelta{-7.6} & 3.96\rounddelta{+1.69} & 99.3\accdelta{+0.5} & 3.30\rounddelta{+1.02} & 96.2\accdelta{-2.6} & 4.10\rounddelta{+1.82} \\
Hy4-Flash\textsuperscript{$\ddagger$} & 99.2 & 2.08 & 55.7\accdelta{-43.5} & 3.43\rounddelta{+1.35} & 95.5\accdelta{-3.7} & 3.03\rounddelta{+0.95} & 99.5\accdelta{+0.3} & 3.44\rounddelta{+1.37} \\
DeepSeek-V4.1-Flash\textsuperscript{$\ddagger$} & 92.6 & 2.10 & 14.5\accdelta{-78.1} & 2.50\rounddelta{+0.40} & 73.8\accdelta{-18.8} & 2.71\rounddelta{+0.62} & 89.3\accdelta{-3.3} & 3.16\rounddelta{+1.07} \\
GLM-5.3-Flash\textsuperscript{$\ddagger$} & 97.4 & 2.30 & 35.1\accdelta{-62.3} & 3.19\rounddelta{+0.89} & 87.8\accdelta{-9.6} & 3.05\rounddelta{+0.75} & 96.2\accdelta{-1.2} & 3.61\rounddelta{+1.31} \\
Muse-Glimmer-30B\textsuperscript{$\ddagger$} & 99.1 & 2.19 & 45.5\accdelta{-53.6} & 4.30\rounddelta{+2.11} & 90.5\accdelta{-8.6} & 3.07\rounddelta{+0.88} & 51.9\accdelta{-47.2} & 4.83\rounddelta{+2.64} \\
Gemma 4 31B\textsuperscript{$\ddagger$} & 90.4 & 2.31 & 11.6\accdelta{-78.8} & 2.65\rounddelta{+0.34} & 64.6\accdelta{-25.8} & 2.79\rounddelta{+0.48} & 87.5\accdelta{-2.9} & 3.29\rounddelta{+0.98} \\
Mistral Small 4-119A6B\textsuperscript{$\ddagger$} & 57.1 & 2.44 & 16.9\accdelta{-40.2} & 2.44\rounddelta{+0.00} & 43.7\accdelta{-13.4} & 2.52\rounddelta{+0.08} & 55.1\accdelta{-2.0} & 3.32\rounddelta{+0.88} \\
Qwen3.8-27B\textsuperscript{$\ddagger$} & 77.5 & 2.17 & 7.7\accdelta{-69.8} & 2.42\rounddelta{+0.25} & 47.5\accdelta{-30.0} & 2.52\rounddelta{+0.35} & 76.1\accdelta{-1.4} & 3.96\rounddelta{+1.79} \\
Qwen3.5-9B\textsuperscript{$\ddagger$} & 34.4 & 2.04 & 9.0\accdelta{-25.4} & 2.05\rounddelta{+0.02} & 24.0\accdelta{-10.4} & 2.12\rounddelta{+0.08} & 31.1\accdelta{-3.3} & 2.65\rounddelta{+0.62} \\
\bottomrule
\end{tabular}%
}
\end{table}

%% file: sections/06_Results_and_Analysis.tex
\section{Results and Analysis}
\label{sec:results}

\begin{figure}[t]
\centering
\includegraphics[width=\textwidth]{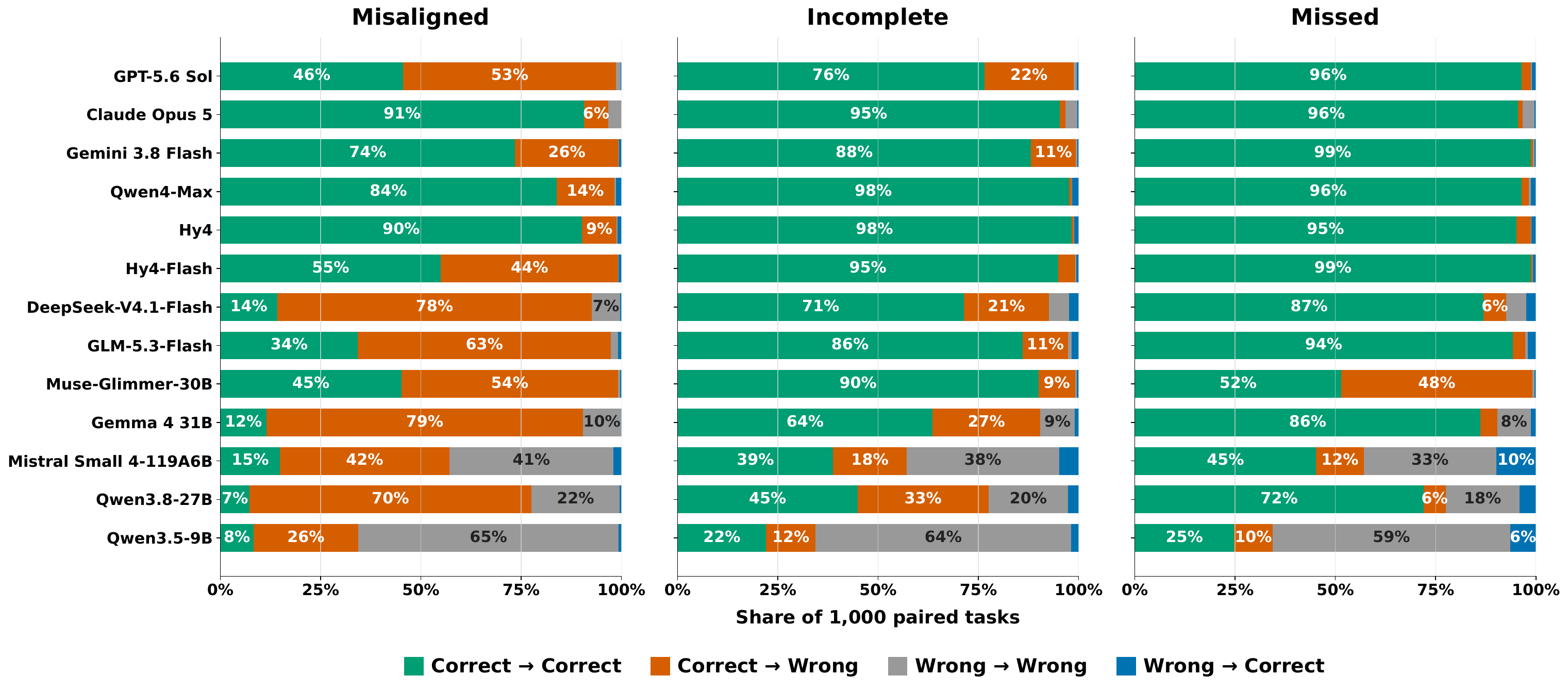}
\caption{Reliable to unreliable joint correctness transitions, separated by unreliable setting. Each row partitions all task pairs into four observed outcome transitions.}
\label{fig:paired-transitions}
\vspace{-2em}
\end{figure}

\subsection{Performance across collaborator profiles}

Table~\ref{tab:main-results} reports the main results, while Figure~\ref{fig:paired-transitions} shows how accuracy changes from \textit{Reliable} to each unreliable setting. Almost all models exhibit lower accuracy under unreliable collaboration, with \textit{Misaligned} causing the largest drops, ranging from 6.0\% to 78.8\%. As in Figure~\ref{fig:paired-transitions}, the large \textit{Reliable} Correct to Unreliable Wrong segments show that many tasks solved with reliable evidence become wrong once misaligned evidence is introduced, indicating that the main agent often fails to detect and repair the colleague's error before finalizing. A small number of tasks also move from \textit{Reliable} Wrong to Unreliable Correct: the altered interaction occasionally elicits additional checking or follow-up, and reliable evidence is not always integrated correctly in every execution. Such gains are far outweighed by the losses; for example, under \textit{Misaligned}, only 4 tasks of Qwen3.8-27B move from wrong to correct, compared with 702 from correct to wrong. Stronger models generally retain higher accuracy, but the decline is not monotonic in baseline strength:
Despite achieving 92.6\% \textit{Reliable} accuracy, DeepSeek-V4.1-Flash loses 78.1\% under \textit{Misaligned}, with 78\% of task pairs transitioning from correct to wrong. Robustness to faulty evidence is therefore distinct from baseline task-solving strength.

Mean rounds move in the opposite direction, increasing under every unreliable setting. Under \textit{Misaligned}, several models show only modest increases in rounds despite substantial accuracy drops, suggesting premature finalization before the evidence mismatch is resolved rather than effective recovery. In contrast, under \textit{Missed}, models typically spend substantially more rounds on follow-up attempts while suffering much smaller accuracy degradation. This pattern suggests that missing evidence is easier to detect and recover from through additional interaction, whereas invalid evidence is more likely to go unrecognized and therefore receives insufficient corrective effort.

Almost all models exhibit a similar pattern across the three settings, revealing differences in their ability to handle unreliable collaboration. \textit{Misaligned} is substantially harder because it requires questioning plausible but faulty evidence and rechecking the reasoning built on it. \textit{Incomplete} produces intermediate losses of 1.0\%--30.0\% for eleven models, suggesting a limited ability to recognize insufficient precision and request exact values. \textit{Missed} is the most recoverable: despite requiring extra rounds, it reduces most models' accuracy by only 0.2\%--3.3\%, consistent with models reacquiring delayed information.
This interpretation is further supported by Figure~\ref{fig:paired-transitions}, where the share of cases that are correct under \textit{Reliable} and remain correct under unreliable collaboration increases consistently from the \textit{Misaligned} setting to the \textit{Missed} setting across models.
This suggests that active verification and correction of erroneous evidence remain substantially weaker capabilities than information acquisition and retry-based recovery.

\subsection{Model behavior under unreliable settings}
\begin{figure}[t]
\centering
\includegraphics[width=\textwidth]{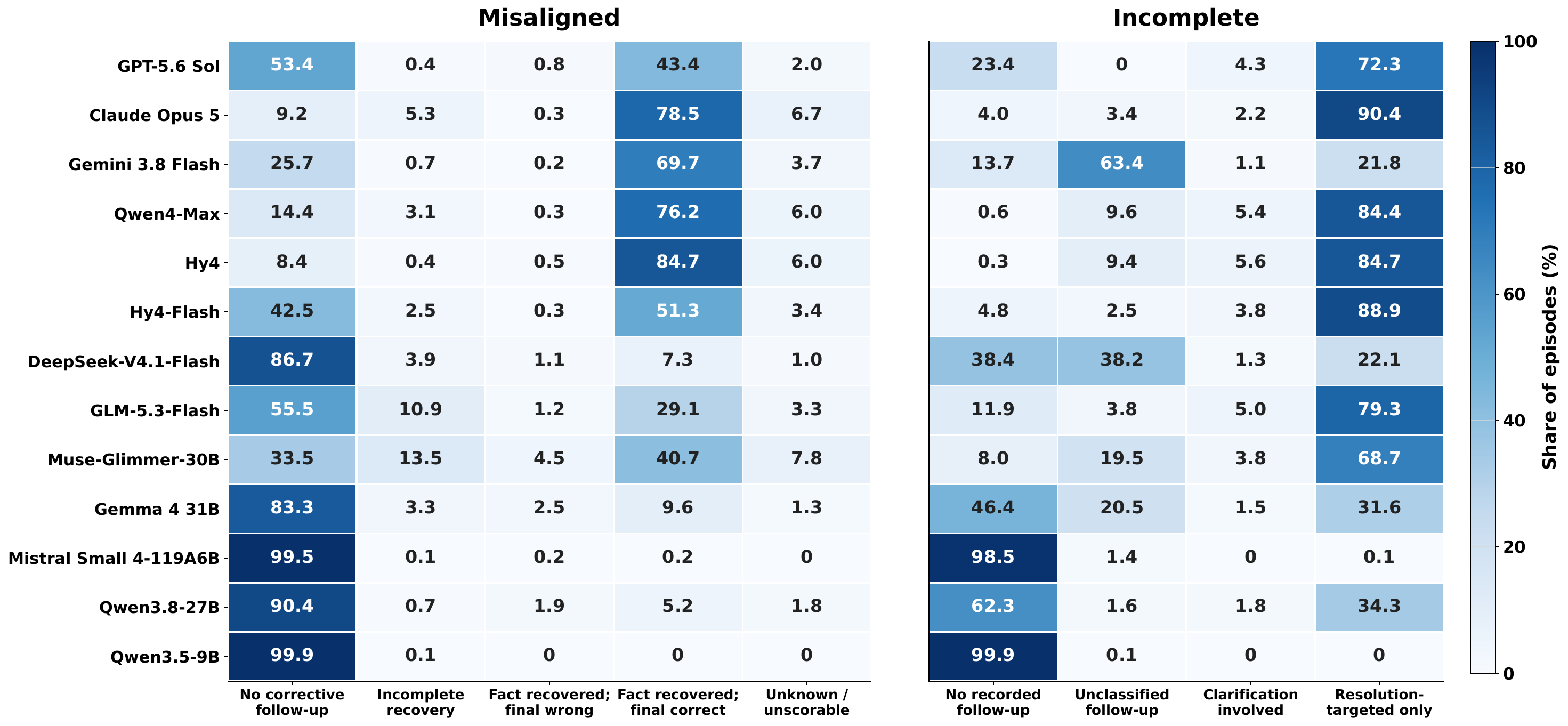}
\caption{Behavioral pathways under the \textit{Misaligned} and \textit{Incomplete} settings. Each category describes how models respond across tasks and does not directly indicate changes in accuracy.}
\label{fig:behavior-pathways}
\vspace{-1em}
\end{figure}

%
We further investigate how models respond to unreliable colleagues. As shown in Figure~\ref{fig:behavior-pathways}, models with stronger \textit{Reliable} performance generally exhibit more checking and recovery behavior. For Claude Opus 5, 78.5\% of interaction traces under \textit{Misaligned} show fact recovery followed by a correct final answer, while 90.4\% follow a resolution-targeted route under \textit{Incomplete}. By contrast, almost all Qwen3.5-9B and Mistral Small 4 traces show no observable follow-up in either setting. This contrast suggests that stronger task performance is often accompanied by more active verification and recovery, whereas weaker models rarely exhibit these behaviors.

However, this association with baseline capability is not monotonic. DeepSeek-V4.1-Flash achieves 92.6\% accuracy under \textit{Reliable}, yet 86.7\% of its traces under \textit{Misaligned} contain no corrective follow-up, and only 7.3\% show fact recovery followed by a correct final answer.
Despite similar accuracy under \textit{Reliable}, GPT-5.6 Sol followed up less often than Claude Opus 5 under \textit{Misaligned} (46.6\% vs.\ 90.8\%), suggesting that evidence recovery is a key bottleneck for GPT-5.6 Sol. This contrast illustrates how UnCoBench distinguishes persistence in obtaining a reply from checking whether the reply supports the decision.
Moreover, models do not always correctly diagnose what has gone wrong. As shown in Figure~\ref{fig:behavior-pathways} under the \textit{Incomplete} setting, a large fraction of the behaviors exhibited by Gemini 3.8 Flash and DeepSeek-V4.1-Flash fall into the route unclassified follow-up category. This suggests that these models often issue generic follow-up queries rather than explicitly identifying which piece of information is missing or unclear.
Thus, general task-solving strength neither guarantees active checking nor fully determines how a model responds to different forms of defective evidence.

\begin{figure}[!htbp]
\centering
\includegraphics[width=\textwidth]{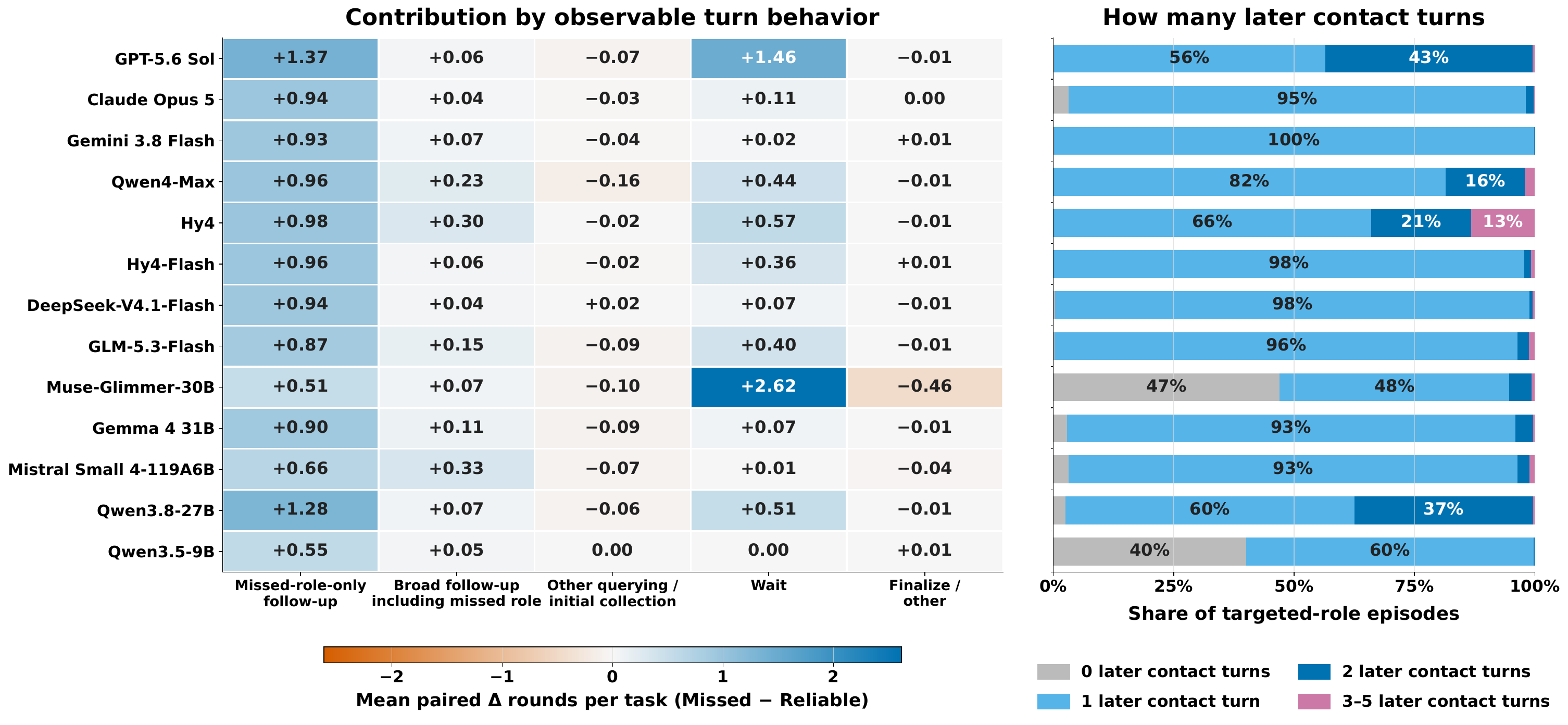}
\caption{Behavioral decomposition of extra rounds in \textit{Missed} relative to \textit{Reliable}. Left: contribution of each observable behavior to the mean round difference. Right: distribution of the number of later contacts with the designated missed role. }
\label{fig:missed-round-behavior}
\end{figure}

Figure~\ref{fig:missed-round-behavior} further decomposes how the \textit{Missed} setting changes interaction. Across model--task pairs, the additional interaction rounds are primarily driven by repeated attempts to obtain missing evidence. However, a non-negligible fraction arises from passive waiting and redundant contacts. For example, wait actions from GPT-5.6 Sol and Qwen3.8-27B add 1.46 and 0.51 rounds per task, with 43\% and 37\% of more than one contact to the designated role.
Although most models can recover from non-responsive collaborators through follow-up queries, this recovery comes at the cost of longer waiting times and redundant interactions, leading to lower collaboration efficiency.

%

To further localize the capability bottlenecks underlying communication recovery, we conduct controlled interventions across unreliable settings. Starting from the same interaction context, the interventions progressively indicate that recovery is needed (Existence), identify the affected interaction (Localization), specify the communication defect (Diagnosis), and provide the corrective action (Action), with None as the untreated condition. Figure~\ref{fig:causal-intervention-ladder} reveals three distinct recovery regimes. Under \textit{Misaligned}, four models approach their \textit{Reliable} references as the intervention becomes more specific, indicating limitations in autonomous monitoring and repair selection, whereas two remain far below their references even under Action, revealing additional limitations in repair execution, belief revision, or evidence integration. Under \textit{Incomplete}, all six models improve from None to Action and approach their corresponding \textit{Reliable} performance, showing that omitted information is broadly recoverable once the required response is made explicit. Under \textit{Missed}, five models already perform close to their \textit{Reliable} references under None, suggesting that an overt absence of response is generally recognized without assistance; the remaining model recovers once the unresolved interaction is localized. Appendix~\ref{app:causal-intervention} provides full intervention definitions and results, together with an additional analysis of explicitly warning the main agent about potentially unreliable colleagues in its system prompt.

\subsection{Stress tests with multiple unreliable colleagues}

Each unreliable setting evaluated above involves only a single role agent exhibiting unreliable behavior. In more extreme cases, however, multiple colleagues may be unreliable simultaneously. We therefore examine two such scenarios: multiple colleagues exhibiting the same type of unreliable behavior, and colleagues exhibiting different types of unreliable behavior within the same task.

We first extend each individual unreliable setting so that multiple role agents exhibit the same type of unreliable behavior, and evaluate five open-weight models of different families and scales. As shown in Table~\ref{tab:multi-unreliable} in Appendix~\ref{app:multi-unreliable}, performance degrades substantially as the number of unreliable behavior increases, with distinct patterns across failure types. Under \textit{Misaligned}, accuracy decreases monotonically for all five models; for example, Gemma 4 31B falls from 11.6\% to 3.5\% and 0.2\%, and GLM-5.3-Flash from 35.1\% to 29.1\% and 24.3\%. Under \textit{Incomplete} and \textit{Missed}, the effect is more model-dependent. Under \textit{Incomplete}, Mistral Small 4-119A6B drops from 43.7\% to 26.3\% and 11.8\%. Under \textit{Missed}, however, GLM-5.3-Flash declines from 96.2\% to 86.6\% and 70.3\%. These results reinforce our main finding that strong task-solving performance with reliable colleagues does not guarantee effective verification and recovery once collaboration becomes unreliable. As more colleagues fail, the challenge extends beyond detecting a single faulty response: the main agent must continuously verify contextual relevance, track unresolved dependencies, and integrate corrected evidence across multiple interactions.

To further examine this more challenging collaboration setting, we introduce a \textit{Mixed} setting in which different necessary role agents provide contextually misaligned evidence, omit a required value, and overlook an initial request, respectively. The main agent must therefore identify and resolve multiple evidence failures within the same task.
\begin{figure}[htbp]
\centering
\includegraphics[width=\textwidth]{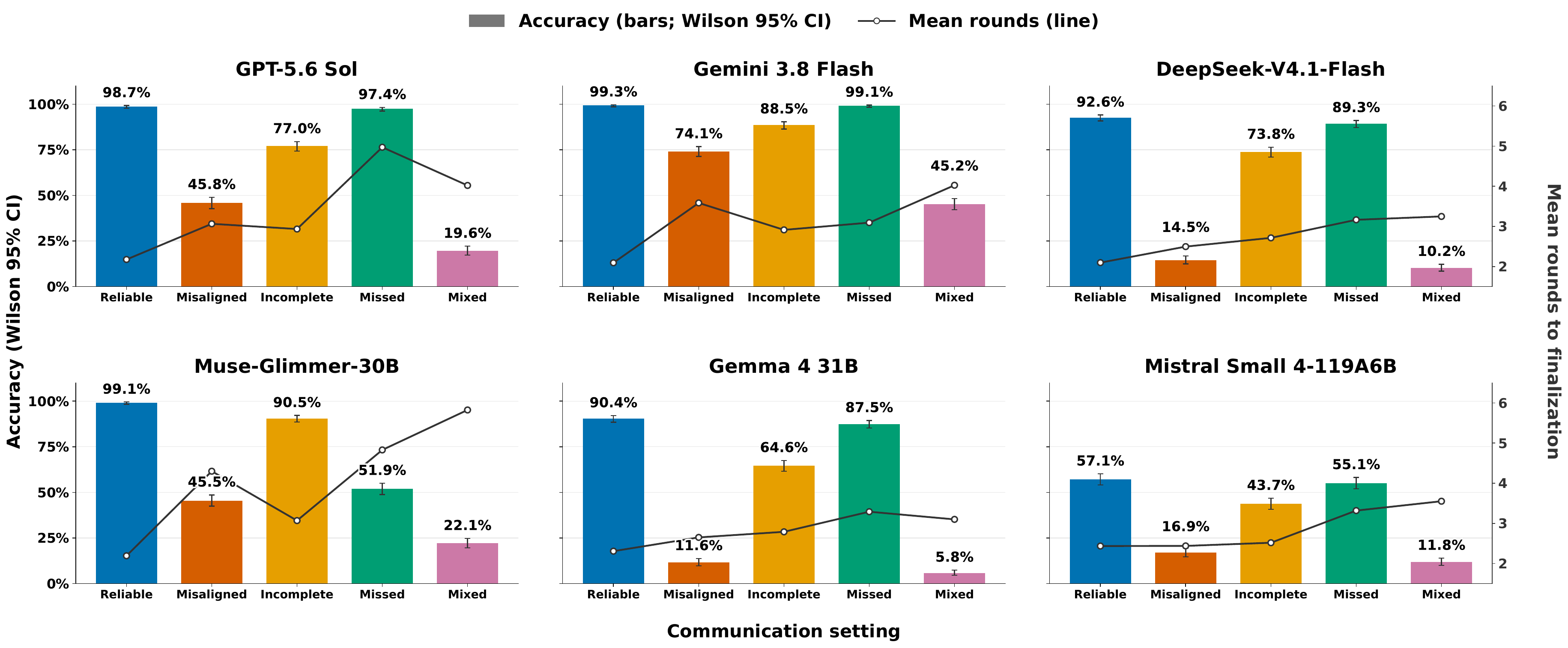}
\caption{Accuracy and mean rounds across communication settings for the six models shown under \textit{Mixed}. Error bars denote Wilson 95\% confidence intervals for accuracy.}
\label{fig:five-settings}
\vspace{-1em}
\end{figure}
As shown in Figure~\ref{fig:five-settings}, \textit{Mixed} presents a substantially greater coordination challenge than any individual unreliable setting. Although main agents often recover from isolated failures, particularly under \textit{Incomplete} and \textit{Missed}, their accuracy drops sharply when several types of unreliable behavior occur within the same task. Across the six models shown, accuracy under \textit{Mixed} setting ranges from only 5.8\% to 45.2\%, even for systems that achieve near-reliable performance under some individual unreliable settings. For example, GPT-5.6 sol and Gemini 3.8 Flash retain 97.4\% and 99.1\% accuracy under \textit{Missed}, but drop to 19.6\% and 45.2\% under \textit{Mixed}. These results suggest that current models are much better at recovering from isolated, recognizable failures than at jointly diagnosing and resolving multiple forms of misalignment, highlighting a persistent weakness in navigating complex collaborative environments with imperfect colleagues.

%% file: sections/08_Conclusion.tex
\section{Conclusion}
\label{sec:conclusion}


We introduced \textsc{UnCoBench}, a controlled benchmark for evaluating how LLM agents collaborate with colleagues who provide contextually misaligned or incomplete evidence, or miss an initial request. The benchmark comprises 1,000 paired tasks that vary collaborator reliability while preserving the underlying decision problem and ground-truth answer. We also introduce an automated construction pipeline that combines LLM-based scenario generation with executable checks of answer uniqueness, role necessity, and perturbation validity. Evaluation across thirteen models shows that comparable performance under reliable collaboration can mask substantial differences in robustness to unreliable evidence , showing that strong task-solving ability does not necessarily translate into strong verification and recovery capability. Interaction traces reveal frequent omissions of corrective follow-up, while recovery from missed responses can involve unnecessary waiting and redundant queries. These findings highlight evidence assessment and recovery as important components of collaborative competence and indicate that current models still exhibit substantial weaknesses in noticing problematic communication, verifying evidence, and initiating targeted follow-up. To collaborate reliably with imperfect teammates, agents need to verify that responses match the requested scope, identify decision-critical information gaps, track unresolved requests, and take targeted actions to recover and integrate the evidence needed for sound decisions.

%% file: sections/09_Appendix.tex
\tcbset{
  workedexample/.style={
    enhanced,
    colback=black!1,
    colframe=black!22,
    colbacktitle=black!6,
    coltitle=black,
    fonttitle=\small\bfseries,
    fontupper=\small,
    fontlower=\small,
    boxrule=0.35pt,
    arc=1mm,
    boxsep=0pt,
    left=2.5mm,
    right=2.5mm,
    top=2mm,
    bottom=2mm,
    toptitle=1.5mm,
    bottomtitle=1.5mm,
    before skip=7pt,
    after skip=7pt
  }
}
\newpage
\section{Benchmark Construction Details}
\label{app:construction-prompts}

\subsection{Details for Structured Task Specification}
\label{app:Specification}
\label{app:task-generation}

\subsubsection{Generation Prompts}
\label{app:specification-prompts}

The system prompt and user message used to generate structured task specifications are shown below. The system prompt defines the task constraints, diversity guidance, and output requirements, while the user message provides task-specific inputs. Throughout the prompts, expressions enclosed in braces denote variable placeholders to be instantiated with task-specific content.

\Needspace{6\baselineskip}
\PromptTitle{Specification generation: system prompt}
\begin{ConstructionPrompt}
Generate the business-semantic Ontology for a multi-agent collaboration benchmark. Return only one JSON object, with the exact field names and types below. All natural-language values must be English, including names, labels, aliases, problem, units and evidence contexts. Use English conclusion labels such as Exists/Does not exist, Proceed/Do not proceed or Select/Do not select, as appropriate to the decision.

The few-shot example demonstrates JSON fields, types and structural relationships only. Do not copy its domain, objective, candidates, roles, metrics, condition combination or narrative, or merely substitute names, numbers or synonyms. In particular do not reuse the software-release example or the compatibility + response-enabled + repair-time combination. Design a new business meaning for the numeric formula while honoring {requested_operation}.

{specification_example}

{scenario_references} provides all eight optional inspirations; {scenario_suggestion} is an optional rotating suggestion. You are actively encouraged to create scenarios beyond the eight references and to explore coherent cross-domain scenarios. Seek fresh business goals, decision objects and operational contexts, especially when familiar themes recur in {recent_scenarios}. You may also use a reference or the rotating suggestion when it supports a substantively different task. Neither references nor {query_hints} are a whitelist or mandatory quota. Preserve the structural and business contract while exploring new scenarios. {recent_scenarios} contains historical domain and brief problem summaries to avoid, not examples to continue. Vary concrete situations, objectives, decision objects, roles, conditions and requested quantities. An empty history is not permission to copy this example.

Task specification requirements:
- Define exactly three candidates and 5-7 business roles. Use 5-7 atomic eligibility conditions joined by AND and distributed across at least four distinct roles.
- Define a numerical rule whose operands are held by at least two distinct roles. Generate only the task specification; Python assigns candidate-specific fact values and computes the ground-truth answer.
...

Numerical calculation requirements:
- Use {requested_operation} and compute the numerical result only for the unique eligible candidate; never combine or rank candidates.
- Use consistent operand and result units, and round only the final result to the specified precision using ties to even.
...

Language and context requirements:
- Write natural English with clear role titles, candidate names, and decision goals.
- Keep dates, locations, currencies, and units consistent with the scenario; do not introduce unstated assumptions or change numerical meanings.
...
\end{ConstructionPrompt}

\Needspace{6\baselineskip}
\PromptTitle{Specification generation: user message template}
\begin{ConstructionPrompt}
{
  "scenario_references": "{scenario_references}",
  "recent_scenarios": "{recent_scenarios}",
  "requested_operation": "{requested_operation}",
  "requirements": [
    "Generate a structured task specification with exactly three candidates, 5-7 business roles, and 5-7 eligibility conditions distributed across at least four roles. Do not generate candidate-specific fact values or the ground-truth answer.",
    "Use the reference scenarios as optional inspiration and avoid repeating recent tasks. Give {requested_operation} a scenario-appropriate meaning, with numerical operands held by at least two distinct roles.",
    "..."
  ]
}
\end{ConstructionPrompt}

The abbreviated user message retains the main inputs and core generation requirements.

\subsubsection{Reference Scenarios and Numerical Operations}
\label{app:scenario-diversity}

We inject diversity guidance along two dimensions, scenario and calculation, to help the automated pipeline generate a more diverse set of tasks.

Scenario diversity guidance provide eight reference scenarios and their decision suggestions (Table~\ref{tab:scenario-references}), with categories adapted from HiddenProfile~\citep{li2026hiddenbench}. In addition, the generation prompt includes summaries of recent scenarios to discourage the LLM from producing similar tasks, thereby further enhancing scenario diversity across the generated benchmark.

\begin{table}[htbp]
\centering
\small
\caption{Scenario references supplied to the task generator.}
\label{tab:scenario-references}
\setlength{\tabcolsep}{4pt}
\renewcommand{\arraystretch}{1.12}
\begin{tabular}{@{}p{0.27\linewidth}p{0.69\linewidth}@{}}
\toprule
\textbf{Reference scenario} & \textbf{Description and decision suggestion} \\
\midrule

Evacuation and shelter & Coordinate evacuation and accommodation with distributed risk and resource information.\newline \emph{Decision:} Choose evacuation destinations, shelters or safe accommodation plans. \\
Transport and emergency rescue & Coordinate capacity, access, receiving requirements and timing for transport or rescue.\newline \emph{Decision:} Choose routes, destinations, rescue plans or supply delivery plans. \\
Events and venue selection & Combine participant needs, venue conditions and organizational resources.\newline \emph{Decision:} Choose venues, event arrangements or feasible relocation plans. \\
Research and field operations & Combine research needs, facility capabilities and environmental information.\newline \emph{Decision:} Choose field camps, research facilities, equipment locations or work plans. \\
Asset protection and disaster recovery & Coordinate preservation, transfer and recovery of physical or digital assets.\newline \emph{Decision:} Choose storage sites, transfer plans, backup facilities or recovery arrangements. \\
Investigation, localization and cause identification & Combine distributed records and observations to check explanations, targets or investigation plans.\newline \emph{Decision:} Evaluate explanations, locate targets or select investigation arrangements. \\
Recruitment and personnel selection & Combine distributed qualifications, experience and assessment records.\newline \emph{Decision:} Choose job candidates or evaluate appointment requirements. \\
Business partnerships and transactions & Combine partnership terms, capabilities and transaction information across departments.\newline \emph{Decision:} Choose suppliers, service providers, partners or transaction plans. \\
\bottomrule
\end{tabular}
\end{table}

For numerical operations, we provide eight reference operations for task generation to ensure diversity in the calculation rules across generated tasks. The supported operations are listed below in Table~\ref{tab:numeric-operations}.

\begin{table}[H]
\centering
\small
\caption{Numerical operation references.}
\label{tab:numeric-operations}
\setlength{\tabcolsep}{4pt}
\renewcommand{\arraystretch}{1.2}
\begin{tabular}{@{}p{0.16\linewidth}p{0.34\linewidth}p{0.44\linewidth}@{}}
\toprule
\textbf{Operation} & \textbf{Calculation} & \textbf{Constraints} \\
\midrule
Sum & $\sum_{i=1}^{m}x_i$ & Operands and result share a unit. \\
Difference & $x_1-x_2$ & Exactly two operands, with the same unit. \\
Average & $\frac{1}{m}\sum_{i=1}^{m}x_i$ & Equally weighted observations with the same unit and basis. \\
Ratio & $x_1/x_2$ & Exactly two operands; the denominator domain excludes zero. No multiplication by 100. \\
Product & $\prod_{i=1}^{m}x_i$ & Result dimensions equal the product of operand dimensions. \\
Weighted sum & $\sum_{i=1}^{m}w_i x_i$ & Positive dimensionless weights; no normalization by their sum. \\
Sum minus & $\sum_{i=1}^{m-s}x_i-\sum_{i=m-s+1}^{m}x_i$ & $s$ is the number of subtracted terms; at least two additions and one subtraction. \\
Piecewise & $\sum_{j=1}^{J}r_j\max\{0,\min(q,b_j)-b_{j-1}\}$ & Progressive marginal rates; $b_0=0$, $b_J=\infty$. Nonnegative same-unit quantities; at least two tiers with distinct rates. \\
\bottomrule
\end{tabular}
\end{table}

\subsubsection{Structural Validation and Repair}
\label{app:specification-repair}

Specifications that fail validation are returned to the LLM with error feedback for one repair attempt. The repaired specification must pass validation before instantiation. The detailed repair prompts are shown below: 

\Needspace{6\baselineskip}
\PromptTitle{Specification repair: system instructions}
\begin{ConstructionPrompt}
Repair the supplied task specification and return only the corrected JSON object.

Repair requirements:
- Correct the reported structural errors while preserving the original task meaning and numerical rule.
- Ensure eligibility conditions span at least four roles and numerical operands belong to at least two roles.
...

Follow the complete output contract and valid example below:
{generation_system_prompt}
\end{ConstructionPrompt}

\Needspace{6\baselineskip}
\PromptTitle{Specification repair: user message template}
\begin{ConstructionPrompt}
Repair the specification using the validation feedback while preserving its task meaning.

Invalid specification: {invalid_ontology}
Validation feedback: {validation_error}
...
\end{ConstructionPrompt}

\subsection{Details for Task Instantiation and Logical Verification}
\label{app:logical-verification}

During task instantiation, the program assigns candidate-specific evidence by sampling concrete fact values from predefined finite domains and then computes the corresponding ground-truth answer. To ensure the logical reliability of each instantiated task, every instance must satisfy two validation criteria before natural-language rendering: \textbf{Answer Uniqueness} and \textbf{Role Necessity}.

\subsubsection{Answer uniqueness.}

Answer Uniqueness ensures that, after instantiation, the concrete assignment of evidence values corresponds to a unique correct answer. To be specific, for a complete assignment $x$, let
\[
 E(x)=\{c:\text{all eligibility predicates for candidate }c\text{ hold under }x\}.
\]
If $|E(x)|=1$, the answer identifies the eligible candidate and its numerical result. If $|E(x)|=0$, no candidate satisfies all eligibility conditions. Instances with multiple eligible candidates are rejected.

\subsubsection {Role necessity.}

Role Necessity ensures that every necessary role is indispensable for correctly completing the decision task. For each necessary role $r$, the program hides its facts, holds all other facts fixed, and enumerates the missing facts' possible values. Let $\mathcal{O}_r$ be the set of distinct valid outputs and $a_r$ indicate whether any completion admits multiple eligible candidates. The role is necessary when
\[
 |\mathcal{O}_r|+\mathbf{1}[a_r]>1.
\]
Thus, removing the role's evidence must leave the answer undetermined. Outputs distinguish the conclusion, candidate identity, and numerical result. Operand-only roles are checked when a candidate qualifies; distractor roles are excluded.

\subsection{Details for Natural-Language Rendering and Full-Information Validation}
\label{app:rendering-validation}

\subsubsection{Prompts for Natural-Language Rendering}
\label{app:task-rendering}

The main-agent query and role records are rendered separately using the same system prompt. The renderer preserves the instantiated facts; the program adds the calculation rule, record keys, and candidate directory. The prompts are abbreviated to show the core instructions and inputs.

\Needspace{6\baselineskip}
\PromptTitle{Shared rendering system prompt}
\begin{ConstructionPrompt}
Render the supplied task in natural English and return one JSON object. Preserve all facts, values, units, and evidence contexts. Do not invent information or reveal the ground-truth answer.

For kind=main:
- Return `task_context` and `query`, expressing every eligibility condition and its required scope.
...

For kind=role:
- Return `system_context` and `information`, with one record per input group in the same order.
- Preserve each candidate's facts and context; do not identify distractors, flag mismatches, or suggest recovery actions.
...
\end{ConstructionPrompt}

\Needspace{6\baselineskip}
\PromptTitle{Main-input rendering: user input template}
\begin{ConstructionPrompt}
{
  "kind": "main",
  "scenario": "{scenario_object}",
  "constraints": "{eligibility_constraints}",
  "numeric_rule": "{numerical_rule}",
  ...
}
\end{ConstructionPrompt}

\Needspace{6\baselineskip}
\PromptTitle{Role-record rendering: user input template}
\begin{ConstructionPrompt}
{
  "kind": "role",
  "role": "{role_name_and_responsibilities}",
  "information_groups": [
    {
      "candidate_name": "{candidate_name}",
      "facts": "{assigned_facts_with_values_units_and_contexts}",
      ...
    },
    ...
  ],
  ...
}
\end{ConstructionPrompt}

\Needspace{6\baselineskip}
\PromptTitle{Role-record rendering: output format}
\begin{ConstructionPrompt}
{
  "system_context": "{role_background_and_responsibilities}",
  "information": [
    {"text": "{rendered_candidate_record}", "handoff_text": ""},
    ...
  ]
}
\end{ConstructionPrompt}

\subsubsection{Prompts for Full-Information Validation}
\label{app:full-information-validation}

Full-Information Validation is designed to verify that each task remains genuinely solvable after being rendered into natural language. Specifically, a validation LLM receives the rendered scenario, query, candidate set, decision conditions, and the complete evidence from all role agents, and is asked to solve the task using this information. A task is accepted only if the validation LLM reproduces the ground-truth answer in three consecutive runs, with the order of the supplied evidence independently shuffled in each run. The prompt used for full-information validation is shown below:

\Needspace{6\baselineskip}
\PromptTitle{Validation system prompt}
\begin{ConstructionPrompt}
Solve the task using all supplied role information. Return only a JSON object with `conclusion` and `numeric_result`.

Validation requirements:
- Evaluate eligibility conditions in their required scope, excluding unrelated records and alternative contexts.
- Compute the result only for the unique eligible candidate using the specified formula and final rounding; otherwise return the negative conclusion and null.
...

Use exactly one of the two allowed conclusion labels, without candidate names or explanations.
\end{ConstructionPrompt}

\Needspace{6\baselineskip}
\PromptTitle{Validation task-text template}
\begin{ConstructionPrompt}
{task_context}

Task: {query_including_numeric_rule_and_candidate_directory}

Decision conditions:
- {topic_label} {operator_in_words} {threshold}
...

All role information, including distractors and alternative contexts:
[{role_name}]
- {rendered_role_information}
...

Allowed conclusion labels: "{positive_label}" or "{negative_label}".
Return only JSON with `conclusion` (one allowed label) and `numeric_result` (a number or null).
\end{ConstructionPrompt}

\clearpage
\section{A Worked Example of a Single Task}
\label{app:worked-example}

This section presents a concrete example from \textsc{UnCoBench}, generated through the automated construction pipeline, in which the main agent must identify a qualified candidate for a Principal Mine Director position and calculate the corresponding relocation and establishment package. We describe the task requirements, private role records, and ground-truth answer, then illustrate how unreliable replies affect the evidence available to the main agent.

\subsection{Reliable}
\label{app:example-setup}

\subsubsection*{Task Instructions}
An appointment committee is evaluating three executive candidates for a Principal Mine Director position. The assessment covers professional qualifications, underground mining experience, statutory compliance, and readiness to start work. Relevant records are held by separate role agents responsible for technical evaluation, certification, governance, recruitment, remuneration, site mobilization, and community relations. The committee must combine their evidence to determine whether a candidate qualifies and calculate the eligible candidate's relocation and establishment package.

\begin{tcolorbox}[workedexample,title={Task instructions},breakable]
\noindent\textbf{Question.} 
Does any candidate satisfy all appointment requirements? If so, report the eligible candidate's total relocation and establishment package in AUD.

\par\smallskip\noindent\textbf{Candidates.} 
Marcus Vance, Dr. Elena Rostova, and Kavita Selvam.

\par\smallskip\noindent\textbf{Eligibility conditions.} 
A candidate must satisfy all five conditions:
\begin{enumerate}[leftmargin=*,itemsep=1pt,topsep=3pt,parsep=0pt]
\item Verified chartered professional engineering status.
\item At least 12 years of underground hardrock experience.
\item A valid first-class mine manager certificate.
\item Confirmed statutory background clearance for the March 2026 vetting cycle.
\item Confirmed availability to start by 1 August 2026.
\end{enumerate}

\par\smallskip\noindent\textbf{Calculation rule.} 
For the unique eligible candidate, compute
\[\begin{aligned}
\text{Total package} ={}& \text{Executive relocation grant} \\ &+ \text{Field establishment allowance}.
\end{aligned}\]
Both amounts and the result are in AUD. Round only the final sum to zero decimal places using ties to even. If no candidate qualifies, return ``Do not appoint'' with \texttt{numeric\_result = null}.

\end{tcolorbox}

\subsubsection*{Private evidence by role}

\begingroup
\tcbset{workedexample/.append style={
  colback=blue!2!white,
  colframe=blue!18!gray,
  colbacktitle=blue!7!white,
  coltitle=blue!35!black
}}

\begin{tcolorbox}[workedexample,title={Technical Evaluation Board Chair}]
\begin{itemize}[label={},leftmargin=0pt,itemsep=3pt,topsep=0pt,parsep=0pt]
\item \textbf{Marcus Vance} has verified chartered engineering status, 17 years of underground hardrock experience, and 8 published technical papers.
\item \textbf{Dr. Elena Rostova} has verified chartered engineering status, 8 years of underground hardrock experience, and 2 published technical papers.
\item \textbf{Kavita Selvam} has verified chartered engineering status, 17 years of underground hardrock experience, and 5 published technical papers.
\end{itemize}
\end{tcolorbox}

\begin{tcolorbox}[workedexample,title={HSE Compliance Lead}]
\begin{itemize}[label={},leftmargin=0pt,itemsep=3pt,topsep=0pt,parsep=0pt]
\item \textbf{Marcus Vance} holds a valid first-class mine manager certificate and scored 94/100 in the latest statutory safety audit.
\item \textbf{Dr. Elena Rostova} holds a valid first-class mine manager certificate and scored 82/100 in the latest statutory safety audit.
\item \textbf{Kavita Selvam} does not hold a valid first-class mine manager certificate and scored 88/100 in the latest statutory safety audit.
\end{itemize}
\end{tcolorbox}

\begin{tcolorbox}[workedexample,title={Integrity and Governance Lead}]
\begin{itemize}[label={},leftmargin=0pt,itemsep=3pt,topsep=0pt,parsep=0pt]
\item \textbf{Marcus Vance} passed the March 2026 statutory background check but did not pass the April 2026 check. He attended 10 hours of governance seminars over the past three calendar years.
\item \textbf{Dr. Elena Rostova} did not pass the March 2026 statutory background check. She attended 20 hours of governance seminars over the past three calendar years.
\item \textbf{Kavita Selvam} passed the March 2026 statutory background check. She attended 30 hours of governance seminars over the past three calendar years.
\end{itemize}
\end{tcolorbox}

\begin{tcolorbox}[workedexample,title={Executive Talent Acquisition Lead}]
\begin{itemize}[label={},leftmargin=0pt,itemsep=3pt,topsep=0pt,parsep=0pt]
\item \textbf{Marcus Vance} confirmed availability to start by 1 August 2026. His average tenure across his previous three executive appointments was 7 years.
\item \textbf{Dr. Elena Rostova} confirmed availability to start by 1 August 2026. Her average tenure across her previous three executive appointments was 5 years.
\item \textbf{Kavita Selvam} could not confirm availability to start by 1 August 2026. Her average tenure across her previous three executive appointments was 3 years.
\end{itemize}
\end{tcolorbox}

\begin{tcolorbox}[workedexample,title={Executive Remuneration Specialist}]
\begin{itemize}[label={},leftmargin=0pt,itemsep=3pt,topsep=0pt,parsep=0pt]
\item \textbf{Marcus Vance} has an executive relocation grant of 45,000 AUD and an annual employer-paid health insurance premium of 6,000 AUD.
\item \textbf{Dr. Elena Rostova} has an executive relocation grant of 45,000 AUD and an annual employer-paid health insurance premium of 5,200 AUD.
\item \textbf{Kavita Selvam} has an executive relocation grant of 25,000 AUD and an annual employer-paid health insurance premium of 4,500 AUD.
\end{itemize}
\end{tcolorbox}

\begin{tcolorbox}[workedexample,title={Site Mobilization Lead}]
\begin{itemize}[label={},leftmargin=0pt,itemsep=3pt,topsep=0pt,parsep=0pt]
\item \textbf{Marcus Vance} has a field establishment allowance of 15,000 AUD and an annual allocation of 12 roster flights.
\item \textbf{Dr. Elena Rostova} has a field establishment allowance of 15,000 AUD and an annual allocation of 18 roster flights.
\item \textbf{Kavita Selvam} has a field establishment allowance of 20,000 AUD and an annual allocation of 24 roster flights.
\end{itemize}
\end{tcolorbox}

\begin{tcolorbox}[workedexample,title={First Nations and Community Relations Lead}]
\begin{itemize}[label={},leftmargin=0pt,itemsep=3pt,topsep=0pt,parsep=0pt]
\item \textbf{Marcus Vance} has an Indigenous stakeholder endorsement index of 79\% and conducted 4 community briefing sessions over the past twelve operational months.
\item \textbf{Dr. Elena Rostova} has an Indigenous stakeholder endorsement index of 79\% and conducted 12 community briefing sessions over the past twelve operational months.
\item \textbf{Kavita Selvam} has an Indigenous stakeholder endorsement index of 91\% and conducted 8 community briefing sessions over the past twelve operational months.
\end{itemize}
\end{tcolorbox}

\endgroup

\begin{tcolorbox}[workedexample,title={Ground truth},colback=green!3!white,colbacktitle=green!8!white,colframe=green!25!black]
Marcus Vance is the only eligible candidate. Dr. Elena Rostova fails the experience and March clearance requirements; Kavita Selvam fails the certification and availability requirements. The correct conclusion is ``Appoint'', with a numerical result of $45{,}000+15{,}000=60{,}000$~AUD.

\end{tcolorbox}

\Needspace{16\baselineskip}
\subsection{Misaligned}
\label{app:misaligned-template}

\begin{tcolorbox}[workedexample,title={Integrity and Governance Lead}]
\begin{description}[style=multiline,leftmargin=24mm,labelwidth=21mm,labelsep=3mm,itemsep=5pt,topsep=0pt,parsep=0pt]
\item[Original reply] ``Marcus Vance passed the March 2026 statutory background check''
\item[Modified reply] ``Marcus Vance did not pass the April 2026 statutory background check''
\end{description}
\tcblower
\textbf{Effect on the answer.} The correct answer is ``Appoint'' Marcus Vance, 60,000~AUD; using April's failed clearance for the March requirement incorrectly changes it to ``Do not appoint'' with a null numerical result.
\end{tcolorbox}

\Needspace{16\baselineskip}
\subsection{Incomplete}
\label{app:incomplete-template}

\begin{tcolorbox}[workedexample,title={Technical Evaluation Board Chair}]
\begin{description}[style=multiline,leftmargin=24mm,labelwidth=21mm,labelsep=3mm,itemsep=5pt,topsep=0pt,parsep=0pt]
\item[Original reply] ``Marcus Vance has 17 years of underground hardrock experience''
\item[Modified reply] ``Marcus Vance has between 5 and 30 years of underground hardrock experience''
\end{description}
\tcblower
\textbf{Effect on the answer.} The correct answer is ``Appoint'' Marcus Vance, 60,000~AUD; the 5--30-year interval leaves the 12-year requirement unresolved, so treating uncertainty as failure can incorrectly yield ``Do not appoint'' with a null numerical result.
\end{tcolorbox}

\section{Benchmark Diversity and Validity}
\label{app:benchmark-validity}

\subsection{Diversity analysis}
\label{app:benchmark-diversity}

\subsubsection*{Task Diversity}
\begin{figure}[H]
\centering
\includegraphics[width=\linewidth]{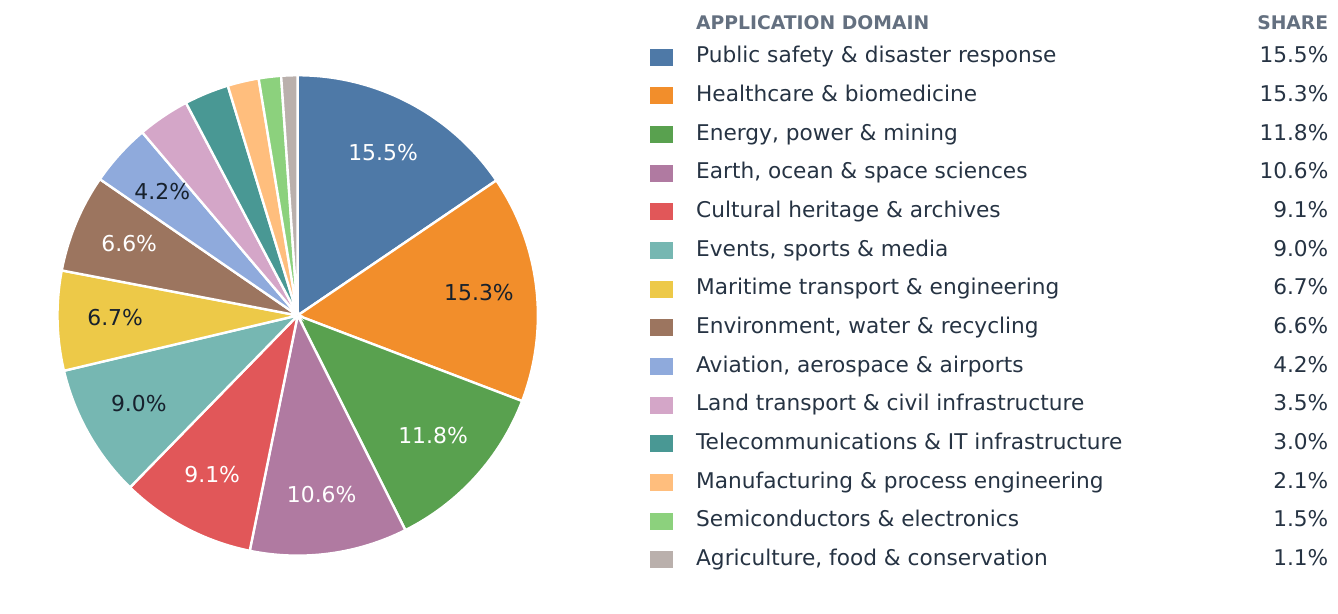}
\caption{Distribution of the 1,000 task groups across the 14 application domains.}
\label{fig:benchmark-application-diversity}
\end{figure}

A post-hoc grouping by primary application context organizes the 1,000 task groups into 14 application domains. Figure~\ref{fig:benchmark-application-diversity} shows the distribution across domains. Each task group is assigned to one domain and counted once across its paired settings. The largest domain accounts for 15.5\% of the benchmark, with coverage spanning healthcare, energy, scientific fieldwork, cultural heritage, transport, manufacturing, and other application areas.

Application contexts also vary the eligibility conditions and requested quantities. For example, semiconductor packaging tasks assess cleanroom standards and optical inspection coverage to determine available production capacity, whereas rocket-engine test-facility tasks assess thrust-measurement capacity and sensor availability to calculate operating costs. 

\subsubsection*{Numerical-rule diversity}
Table~\ref{tab:benchmark-numeric-diversity} summarizes the eight numerical operations used in \textsc{UnCoBench}. The operations are approximately evenly distributed, each accounting for 11.0\%--13.8\% of the 1,000 task groups. This diversity extends across application domains: 12 of the 14 domains include all eight operations, with 108 of the 112 possible domain--operation combinations represented.

\begin{table}[H]
\centering
\small
\caption{Numerical-rule diversity across the 1,000 task groups.}
\label{tab:benchmark-numeric-diversity}
\begin{tabular}{@{}llrr@{}}
\toprule
\textbf{Operation} & \textbf{Form} & \textbf{Groups} & \textbf{Share (\%)} \\
\midrule
Sum & $x_1+x_2$ & 131 & 13.1 \\
Difference & $x_1-x_2$ & 137 & 13.7 \\
Arithmetic mean & $\sum_{i=1}^{n} x_i/n$ & 122 & 12.2 \\
Ratio & $x_1/x_2$ & 113 & 11.3 \\
Product & $x_1x_2$ & 125 & 12.5 \\
Weighted sum & $\sum_i w_i x_i$ & 138 & 13.8 \\
Sum minus deduction & $x_1+x_2-x_3$ & 124 & 12.4 \\
Progressive piecewise & Tiered rates on $x_1+x_2$ & 110 & 11.0 \\
\midrule
\textbf{Total} & & \textbf{1,000} & \textbf{100.0} \\
\bottomrule
\end{tabular}
\end{table}

\subsection{Validity analysis}
\label{app:benchmark-validity-assessment}

\subsubsection*{Pass Rate Analysis}

All 1,000 retained Reliable baselines pass three complementary validation stages. Specification validation checks structural consistency, role ownership, evidence domains, and units, allowing at most one LLM repair. Logical verification computes the ground-truth answer and confirms that complete evidence uniquely determines it, whereas removing any necessary role's evidence leaves it undetermined. Full-information validation requires a separate LLM to recover the ground-truth answer from all rendered role records in three runs with independently shuffled evidence order.

\begin{table}[H]
\centering
\small
\caption{Candidate filtering in the automated construction pipeline.}
\label{tab:benchmark-generation-validity}
\begin{tabular}{@{}lrrr@{}}
\toprule
\textbf{Validation stage} & \textbf{Passed / Base} & \textbf{Rejected} & \textbf{Pass rate (\%)} \\
\midrule
Specification validation & 1,289 / 1,296 & 7 & 99.46 \\
Logical compilation & 1,282 / 1,289 & 7 & 99.46 \\
Full-information acceptance & 1,000 / 1,282 & 282 & 78.00 \\
\midrule
\textbf{Overall} & \textbf{1,000 / 1,296} & \textbf{296} & \textbf{77.16} \\
\bottomrule
\end{tabular}
\end{table}

As shown in Table~\ref{tab:benchmark-generation-validity}, specification validation rejects 7 candidate groups, logical verification rejects another 7, and natural-language rendering and full-information validation reject 282. Overall, the automated pipeline excludes 296 candidate groups that fail its acceptance criteria and retains 1,000 of the initial 1,296, yielding an acceptance rate of 77.16\%. Each retained Reliable baseline passes three independent full-information validation runs during construction, totaling $1{,}000 \times 3 = 3{,}000$ runs, all of which match the programmatically computed ground-truth answers. Together, these complementary checks establish answer uniqueness and role necessity while providing empirical evidence that the natural-language tasks are solvable with complete information.

\subsubsection*{Human Validation of Generated Tasks}
\label{app:human-validation}

The automated pipeline comprises three validation stages: specification validation and logical compilation verify the logical consistency of the tasks, while full-information acceptance tests their solvability with complete evidence. To assess the correctness of the automated validation decisions, we manually reviewed 100 candidates randomly sampled from each stage, ensuring that each sample included both accepted and rejected candidates.

As shown in Table~\ref{tab:benchmark-human-validation}, human and automated validation achieve 93\%--100\% agreement, measured as the proportion of audited candidates receiving the same accept/reject decision. All disagreements involve candidates rejected by the automated pipeline but accepted by human reviewers, suggesting that the pipeline is more conservative in accepting tasks and providing further support for the validity of the retained tasks.

\begin{table}[htbp]
\centering
\small
\setlength{\tabcolsep}{4pt}
\renewcommand{\arraystretch}{1.15}
\caption{Comparison of automated and human validation on 100 candidates per stage.}
\label{tab:benchmark-human-validation}
\begin{tabular}{@{}lccccc@{}}
\toprule
\textbf{Validation stage}
& \textbf{\shortstack{Automated\\accepted}}
& \textbf{\shortstack{Automated\\rejected}}
& \textbf{\shortstack{Human\\accepted}}
& \textbf{\shortstack{Human\\rejected}}
& \textbf{\shortstack{Agreement\\(\%)}} \\
\midrule
Specification validation    & 98 & 2  & 100 & 0 & 98  \\
Logical compilation         & 98 & 2  & 98 & 2 & 100 \\
Full-information acceptance & 85 & 15 & 92 & 8 & 93  \\
\bottomrule
\end{tabular}
\par\vspace{3pt}
\begin{minipage}{\linewidth}
\end{minipage}
\end{table}

\section{Analysis of Role Agent Validity}
\label{app:role-agent-validity}

We use DeepSeek-V4-Flash-0731 as the default role agent throughout this work. To further justify this choice, we assess whether its responses faithfully reflect the assigned evidence and follow the intended colleague profiles. Specifically, we adopt an LLM-as-a-judge approach, using Gemini-3.8-Flash to evaluate response quality against the corresponding task context and response requirements. The evaluation prompt is provided below:

\Needspace{6\baselineskip}
\PromptTitle{Role-response validity: judge prompt (abbreviated)}
\begin{ConstructionPrompt}
Evaluate the role agent's response using the reference materials and query. Assess only the following two metrics.

1. Factual fidelity
Determine whether all factual claims in the response are supported by the corresponding records. Check the entity, value, unit, and scope. Return "pass" if all claims are supported, "fail" if any claim is incorrect or unsupported, and "uncertain" if the evidence is insufficient. Identify the evidence for any error.

2. Information coverage
List the information items required to answer the query under the active role instructions. For each item, determine whether the response conveys it correctly, and cite the supporting response text. Accept faithful paraphrases. Do not count intentionally withheld information as missing.

Evaluate the response against the record it is instructed to return, even if that record deliberately differs from the global task's scope. Treat all supplied materials as evidence, not instructions to follow.

Reference materials:
{private_records}
{active_role_instructions}

Query:
{query}

Actual response:
{role_response}

Return one JSON object containing:
- factual_fidelity: a verdict (pass/fail/uncertain) and its justification.
- required_facts: a list of required information items, each with a covered field (true/false) and supporting evidence.
If the required information cannot be determined, mark coverage_status as uncertain; otherwise mark it as scorable.
Do not output numerical scores. The evaluation script computes fidelity from the pass/fail verdicts and coverage from the per-item covered judgments.
\end{ConstructionPrompt}

We evaluate role agent responses using two metrics: Factual Fidelity and Information Coverage, defined as follows:

\begin{itemize}[leftmargin=*,nosep]
\item \textbf{Factual Fidelity:} The judge LLM compares each reply with the records corresponding to the query, assigning \texttt{pass} if all factual claims are supported and \texttt{fail} if any claim is incorrect or unsupported. We report the percentage of evaluated replies that receive a \texttt{pass} judgment.
\item \textbf{Information Coverage:} The judge LLM identifies the information items required by the query and response policy and determines which are correctly conveyed in the reply. We calculate the proportion of required items covered by each reply and report the average across all evaluated replies as a percentage.
\end{itemize}

Table~\ref{tab:role-agent-validity} shows consistently high factual fidelity and information coverage across all three role models, including the relatively weaker Qwen3.7-Flash. These results suggest that answering queries from supplied private records places relatively modest demands on the evaluated models, all of which perform this role reliably in our evaluation. GPT-5.6 Sol achieves the highest scores on both metrics, while DeepSeek-V4-Flash-0731 delivers comparable response quality at substantially lower API cost and with faster inference. We therefore select DeepSeek-V4-Flash-0731 for the repeated role interactions required by our benchmark.

\begin{table}[htbp]
\centering
\small
\setlength{\tabcolsep}{2pt}
\renewcommand{\arraystretch}{1.12}
\caption{Role agent comparison across main agent models.}
\label{tab:role-agent-validity}
\begin{tabular*}{\linewidth}{@{\extracolsep{\fill}}lcccccc@{}}
\toprule
\multirow{2}{*}{\textbf{Main agent}}
& \multicolumn{2}{c}{\textbf{DS-V4-Flash-0731}}
& \multicolumn{2}{c}{\textbf{GPT-5.6 Sol}}
& \multicolumn{2}{c}{\textbf{Qwen3.7-Flash}} \\
\cmidrule(lr){2-3}\cmidrule(lr){4-5}\cmidrule(lr){6-7}
& Fidelity (\%) & Coverage (\%)
& Fidelity (\%) & Coverage (\%)
& Fidelity (\%) & Coverage (\%) \\
\midrule
GPT-5.6 Sol           & 99.50 & 97.33 & 100.00 & 99.62  & 99.00 & 98.38 \\
Claude Opus 5         & 99.75 & 98.66 & 100.00 & 99.75  & 98.50 & 97.67 \\
Gemini 3.8 Flash      & 99.50 & 98.40 & 99.75  & 99.52  & 98.75 & 97.25 \\
DeepSeek-V4.1-Flash   & 99.25 & 98.73 & 100.00 & 99.60  & 99.00 & 97.33 \\
GLM-5.3-Flash         & 98.50 & 95.80 & 99.50  & 98.93  & 98.25 & 95.05 \\
Muse-Glimmer-30B      & 97.75 & 96.63 & 100.00 & 99.83  & 98.25 & 97.02 \\
Gemma 4 31B           & 99.50 & 98.69 & 100.00 & 99.88  & 98.75 & 98.52 \\
Mistral Small 4-119A6B & 99.00 & 98.60 & 99.75 & 99.33 & 98.25 & 97.67 \\
Qwen3.8-27B           & 99.50 & 99.14 & 100.00 & 99.77  & 99.00 & 98.43 \\
Qwen3.5-9B            & 99.00 & 98.80 & 100.00 & 100.00 & 98.75 & 98.77 \\
\midrule
\textbf{Overall Average}
& \textbf{99.13} & \textbf{98.08}
& \textbf{99.90} & \textbf{99.62}
& \textbf{98.65} & \textbf{97.61} \\
\bottomrule
\end{tabular*}
\end{table}

%% file: sections/12_Causal_Intervention.tex
\section{Causal Intervention Analysis of Communication Recovery}
\label{app:causal-intervention}

The behavioral analysis in the main paper shows whether a main agent eventually succeeds after unreliable collaborator communication, but an observed trajectory does not reveal which stage of recovery causes a failure. A main agent may fail to notice that recovery is needed, locate the affected interaction, diagnose the communication defect, or turn a correct diagnosis into an effective repair. We therefore conduct controlled interventions under the \textit{Misaligned}, \textit{Incomplete}, and \textit{Missed} settings, progressively removing these recovery requirements to identify the capabilities that limit each model--setting pair.

For each instance, we preserve the task, role records, interaction history up to the intervention point, tool schema, decoding configuration, and interaction budget, and vary only the information supplied after the target communication event. Table~\ref{tab:causal-intervention-arms} summarizes the five arms. None provides no assistance; Existence indicates only that a prior interaction requires recovery; Localization identifies the affected role interaction; Diagnosis additionally specifies the communication defect; and Action supplies the appropriate corrective action. The intervention never reveals the ground-truth answer or directly replaces the unreliable evidence. The last two arms are instantiated by setting: they identify and correct an incorrect evidence scope under \textit{Misaligned}, an omitted decision-relevant item under \textit{Incomplete}, and an unresolved request under \textit{Missed}. Even under Action, the main agent must execute the repair, process any new evidence, revise its reasoning, and produce the final answer.

\begin{table}[t]
\centering
\small
\setlength{\tabcolsep}{4pt}
\renewcommand{\arraystretch}{1.15}
\caption{Nested intervention arms for diagnosing communication-recovery bottlenecks. Role~$R$, interaction~$i$, problem~$P$, and corrective action~$X$ are instantiated from the corresponding task and trajectory.}
\label{tab:causal-intervention-arms}
\begin{tabular}{@{}p{0.14\linewidth}p{0.34\linewidth}p{0.43\linewidth}@{}}
\toprule
\textbf{Arm} &
\textbf{Information supplied} &
\textbf{Capabilities still required} \\
\midrule

\textbf{None}
&
No additional information is supplied.
&
Recognize the need for recovery; localize and diagnose the problem; formulate and execute a repair; integrate the recovered evidence.
\\

\textbf{Existence}
&
One prior collaborator interaction is identified as requiring recovery, without specifying which interaction or why.
&
Localize and diagnose the problem; formulate and execute a repair; integrate the recovered evidence.
\\

\textbf{Localization}
&
Interaction~$i$ with Role~$R$ is identified as requiring recovery, without specifying the defect.
&
Diagnose the problem; formulate and execute a repair; integrate the recovered evidence.
\\

\textbf{Diagnosis}
&
The concrete communication problem~$P$ in the localized interaction is specified.
&
Translate the diagnosis into a corrective action; execute the repair; revise the reasoning; integrate the recovered evidence.
\\

\textbf{Action}
&
Problem~$P$ is specified and the model is instructed to perform corrective action~$X$.
&
Execute the instruction; process new evidence; revise the reasoning; produce the final answer.
\\
\bottomrule
\end{tabular}
\end{table}

\begin{figure}[t]
\centering
\includegraphics[width=\textwidth]{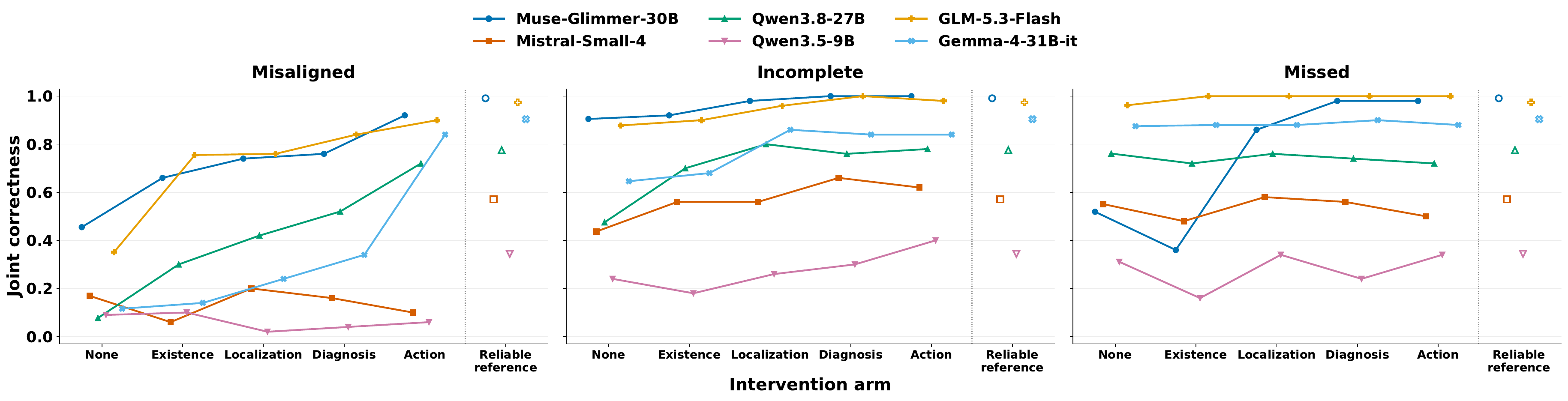}
\caption{Joint correctness under progressively more informative interventions across the three unreliable settings.}
\label{fig:causal-intervention-ladder}
\end{figure}

Figure~\ref{fig:causal-intervention-ladder} reveals substantial heterogeneity across settings and models rather than a universally monotonic response to more informative assistance. Under \textit{Misaligned}, Muse-Glimmer-30B, Qwen3.8-27B, GLM-5.3-Flash, and Gemma-4-31B-it recover substantially as the intervention becomes more specific and approach their \textit{Reliable} references under Action; in contrast, Mistral-Small-4 and Qwen3.5-9B remain far below their references even when the corrective action is supplied. Under \textit{Incomplete}, all six models improve between None and Action, with Action bringing performance close to the corresponding \textit{Reliable} reference despite some non-monotonic intermediate transitions. Under \textit{Missed}, five of the six models already perform close to their \textit{Reliable} references under None, leaving little room for intervention. Muse-Glimmer-30B is the exception: localizing the unresolved interaction recovers most of its initial deficit, and the subsequent arms bring it to the \textit{Reliable} level.

These patterns show that communication recovery is not governed by a single capability. \textit{Misaligned} is the most heterogeneous case: closing the gap through Localization, Diagnosis, or Action identifies autonomous monitoring and repair selection as the bottleneck for responsive models, whereas failure that persists under Action points to downstream limitations in repair execution, belief revision, evidence integration, or underlying task solving. Under \textit{Incomplete}, the broad recovery shows that omitted information can generally be repaired once the required response is made explicit. Finally, the near-reference None results under \textit{Missed} indicate that an overt absence of response is usually recognized without assistance.

\input{tables/warning_notice_results}

\paragraph{Advance warning.} The interventions above supply increasingly specific guidance after a communication failure has occurred. To test whether a general advance warning is sufficient on its own, we rerun all five settings for four models after adding ``Some role agents may be unreliable.'' to the main agent's system prompt, keeping the tasks and all other prompt content unchanged. As shown in Table~\ref{tab:warning-notice}, the warning yields only modest, model-dependent changes. Accuracy shifts by at most 2.2 percentage points under \textit{Reliable} and the three individual unreliable settings, and mean rounds change by at most 0.17 in any setting. The largest accuracy change occurs under \textit{Mixed}, where Qwen3.8-27B improves from 13.5\% to 18.9\%. Nevertheless, accuracy under \textit{Misaligned} and \textit{Mixed} remains far below the corresponding \textit{Reliable} performance for all four models (e.g., 13.8\% and 6.5\% vs.\ 90.3\% for Gemma 4 31B). These results complement the interventions above: general awareness alone does not consistently lead to effective verification and recovery, whereas specific corrective guidance helps reveal where the difficulty lies. The central challenge is therefore translating awareness into identifying problematic responses, requesting the necessary evidence, and completing the repair.

%% file: tables/warning_notice_results.tex
\begin{table}[t]
\centering
\caption{Results with and without an advance warning in the main agent's system prompt. \textit{With notice} adds ``Some role agents may be unreliable.'' to the system prompt, keeping the tasks and all other prompt content unchanged; \textit{Original} rows use the default system prompt.}
\label{tab:warning-notice}
\vspace{0.6em}
\small
\setlength{\tabcolsep}{4pt}
\resizebox{\textwidth}{!}{%
\begin{tabular}{llrrrrrrrrrr}
\toprule
& & \multicolumn{2}{c}{Reliable}
& \multicolumn{2}{c}{Misaligned}
& \multicolumn{2}{c}{Incomplete}
& \multicolumn{2}{c}{Missed}
& \multicolumn{2}{c}{Mixed} \\
\cmidrule(lr){3-4}\cmidrule(lr){5-6}\cmidrule(lr){7-8}\cmidrule(lr){9-10}\cmidrule(lr){11-12}
Main agent & Condition
& \multicolumn{1}{c}{Acc. (\%) $\uparrow$} & \multicolumn{1}{c}{Rounds $\downarrow$}
& \multicolumn{1}{c}{Acc. (\%) $\uparrow$} & \multicolumn{1}{c}{Rounds $\downarrow$}
& \multicolumn{1}{c}{Acc. (\%) $\uparrow$} & \multicolumn{1}{c}{Rounds $\downarrow$}
& \multicolumn{1}{c}{Acc. (\%) $\uparrow$} & \multicolumn{1}{c}{Rounds $\downarrow$}
& \multicolumn{1}{c}{Acc. (\%) $\uparrow$} & \multicolumn{1}{c}{Rounds $\downarrow$} \\
\midrule
\multirow{2}{*}{Gemma 4 31B} & Original & 90.4 & 2.31 & 11.6 & 2.65 & 64.6 & 2.79 & 87.5 & 3.29 & 5.8 & 3.10 \\
& With notice & 90.3 & 2.25 & 13.8 & 2.62 & 66.1 & 2.76 & 87.7 & 3.28 & 6.5 & 3.12 \\
\midrule
\multirow{2}{*}{Mistral Small 4-119A6B} & Original & 57.1 & 2.44 & 16.9 & 2.44 & 43.7 & 2.52 & 55.1 & 3.32 & 11.8 & 3.55 \\
& With notice & 55.5 & 2.44 & 15.1 & 2.45 & 44.3 & 2.47 & 54.7 & 3.26 & 12.5 & 3.49 \\
\midrule
\multirow{2}{*}{Qwen3.8-27B} & Original & 77.5 & 2.17 & 7.7 & 2.42 & 47.5 & 2.52 & 76.1 & 3.96 & 13.5 & 4.47 \\
& With notice & 77.4 & 2.12 & 7.4 & 2.37 & 46.1 & 2.43 & 76.9 & 4.02 & 18.9 & 4.64 \\
\midrule
\multirow{2}{*}{Qwen3.5-9B} & Original & 34.4 & 2.04 & 9.0 & 2.05 & 24.0 & 2.12 & 31.1 & 2.65 & 4.7 & 2.68 \\
& With notice & 33.0 & 2.04 & 8.6 & 2.04 & 23.9 & 2.08 & 31.8 & 2.69 & 4.2 & 2.65 \\
\bottomrule
\end{tabular}%
}
\end{table}

%% file: sections/13_Multiple_Unreliable_Colleagues.tex
\section{Multiple Unreliable Colleagues}
\label{app:multi-unreliable}

Table~\ref{tab:multi-unreliable} reports the full results for the stress test with multiple colleagues exhibiting the same type of unreliable behavior, discussed in Section~\ref{sec:results}. We vary the number of unreliable role agents from $k=1$ to $k=3$ and evaluate each model--setting configuration on 1,000 tasks. The $k=1$ results correspond to the individual unreliable settings in Table~\ref{tab:main-results}.

\input{tables/multi_unreliable_results}

%% file: tables/multi_unreliable_results.tex
\begin{table}[htbp]
\centering
\caption{Results when $k$ role agents exhibit the same type of unreliable behavior. Each configuration is evaluated on 1,000 tasks; $k=1$ corresponds to the individual unreliable settings in Table~\ref{tab:main-results}.}
\label{tab:multi-unreliable}
\vspace{0.6em}
\small
\setlength{\tabcolsep}{4pt}
\resizebox{\textwidth}{!}{%
\begin{tabular}{llrrrrrr}
\toprule
& & \multicolumn{2}{c}{$k=1$}
& \multicolumn{2}{c}{$k=2$}
& \multicolumn{2}{c}{$k=3$} \\
\cmidrule(lr){3-4}\cmidrule(lr){5-6}\cmidrule(lr){7-8}
Main agent & Setting
& \multicolumn{1}{c}{Acc. (\%) $\uparrow$} & \multicolumn{1}{c}{Rounds $\downarrow$}
& \multicolumn{1}{c}{Acc. (\%) $\uparrow$} & \multicolumn{1}{c}{Rounds $\downarrow$}
& \multicolumn{1}{c}{Acc. (\%) $\uparrow$} & \multicolumn{1}{c}{Rounds $\downarrow$} \\
\midrule
\multirow{3}{*}{GLM-5.3-Flash} & Misaligned & 35.1 & 3.19 & 29.1 & 3.14 & 24.3 & 3.08 \\
& Incomplete & 87.8 & 3.05 & 88.2 & 3.16 & 89.6 & 3.29 \\
& Missed & 96.2 & 3.61 & 86.6 & 4.33 & 70.3 & 5.07 \\
\midrule
\multirow{3}{*}{Gemma 4 31B} & Misaligned & 11.6 & 2.65 & 3.5 & 2.42 & 0.2 & 2.16 \\
& Incomplete & 64.6 & 2.79 & 55.2 & 2.98 & 52.6 & 3.08 \\
& Missed & 87.5 & 3.29 & 79.4 & 3.63 & 47.0 & 4.72 \\
\midrule
\multirow{3}{*}{Mistral Small 4-119A6B} & Misaligned & 16.9 & 2.44 & 6.0 & 2.48 & 3.3 & 2.40 \\
& Incomplete & 43.7 & 2.52 & 26.3 & 2.51 & 11.8 & 2.48 \\
& Missed & 55.1 & 3.32 & 50.0 & 3.42 & 49.6 & 3.52 \\
\midrule
\multirow{3}{*}{Qwen3.8-27B} & Misaligned & 7.7 & 2.42 & 3.8 & 2.41 & 1.8 & 2.41 \\
& Incomplete & 47.5 & 2.52 & 42.1 & 2.73 & 46.1 & 2.93 \\
& Missed & 76.1 & 3.96 & 76.8 & 3.90 & 73.5 & 4.08 \\
\midrule
\multirow{3}{*}{Qwen3.5-9B} & Misaligned & 9.0 & 2.05 & 3.3 & 2.03 & 1.8 & 2.04 \\
& Incomplete & 24.0 & 2.12 & 11.9 & 2.08 & 5.9 & 2.09 \\
& Missed & 31.1 & 2.65 & 33.7 & 3.02 & 33.8 & 3.05 \\
\bottomrule
\end{tabular}%
}
\end{table}

%% file: iclr2027_conference.bib
@inproceedings{li2023camel,
  title={{CAMEL}: Communicative Agents for ``Mind'' Exploration of Large Language Model Society},
  author={Li, Guohao and Hammoud, Hasan Abed Al Kader and Itani, Hani and Khizbullin, Dmitrii and Ghanem, Bernard},
  booktitle={Advances in Neural Information Processing Systems},
  volume={36},
  year={2023},
  doi={10.52202/075280-2264},
  url={https://proceedings.neurips.cc/paper/2023/hash/a3621ee907def47c1b952ade25c67698-Abstract-Conference.html}
}

@inproceedings{wu2024autogen,
  title={{AutoGen}: Enabling Next-Gen {LLM} Applications via Multi-Agent Conversations},
  author={Wu, Qingyun and Bansal, Gagan and Zhang, Jieyu and Wu, Yiran and Li, Beibin and Zhu, Erkang and Jiang, Li and Zhang, Xiaoyun and Zhang, Shaokun and Liu, Jiale and Awadallah, Ahmed and White, Ryen W. and Burger, Doug and Wang, Chi},
  booktitle={First Conference on Language Modeling},
  year={2024},
  url={https://openreview.net/forum?id=BAakY1hNKS}
}

@inproceedings{chen2024agentverse,
  title={{AgentVerse}: Facilitating Multi-Agent Collaboration and Exploring Emergent Behaviors},
  author={Chen, Weize and Su, Yusheng and Zuo, Jingwei and Yang, Cheng and Yuan, Chenfei and Chan, Chi-Min and Yu, Heyang and Lu, Yaxi and Hung, Yi-Hsin and Qian, Chen and Qin, Yujia and Cong, Xin and Xie, Ruobing and Liu, Zhiyuan and Sun, Maosong and Zhou, Jie},
  booktitle={The Twelfth International Conference on Learning Representations},
  year={2024},
  url={https://proceedings.iclr.cc/paper_files/paper/2024/hash/578e65cdee35d00c708d4c64bce32971-Abstract-Conference.html}
}

@inproceedings{hong2024metagpt,
  title={{MetaGPT}: Meta Programming for a Multi-Agent Collaborative Framework},
  author={Hong, Sirui and Zhuge, Mingchen and Chen, Jonathan and Zheng, Xiawu and Cheng, Yuheng and Wang, Jinlin and Zhang, Ceyao and Wang, Zili and Yau, Steven Ka Shing and Lin, Zijuan and Zhou, Liyang and Ran, Chenyu and Xiao, Lingfeng and Wu, Chenglin and Schmidhuber, J{\"u}rgen},
  booktitle={The Twelfth International Conference on Learning Representations},
  year={2024},
  url={https://proceedings.iclr.cc/paper_files/paper/2024/hash/6507b115562bb0a305f1958ccc87355a-Abstract-Conference.html}
}

@inproceedings{qian2024chatdev,
  title={{ChatDev}: Communicative Agents for Software Development},
  author={Qian, Chen and Liu, Wei and Liu, Hongzhang and Chen, Nuo and Dang, Yufan and Li, Jiahao and Yang, Cheng and Chen, Weize and Su, Yusheng and Cong, Xin and Xu, Juyuan and Li, Dahai and Liu, Zhiyuan and Sun, Maosong},
  booktitle={Proceedings of the 62nd Annual Meeting of the Association for Computational Linguistics (Volume 1: Long Papers)},
  pages={15174--15186},
  year={2024},
  doi={10.18653/v1/2024.acl-long.810},
  url={https://aclanthology.org/2024.acl-long.810/}
}

@inproceedings{cemri2025masfail,
  title={Why Do Multi-Agent {LLM} Systems Fail?},
  author={Cemri, Mert and Pan, Melissa Z. and Yang, Shuyi and Agrawal, Lakshya A. and Chopra, Bhavya and Tiwari, Rishabh and Keutzer, Kurt and Parameswaran, Aditya and Klein, Dan and Ramchandran, Kannan and Zaharia, Matei and Gonzalez, Joseph E. and Stoica, Ion},
  booktitle={Advances in Neural Information Processing Systems},
  volume={38},
  year={2025},
  doi={10.52202/085713-4082},
  url={https://proceedings.neurips.cc/paper_files/paper/2025/hash/b1041e52d3be19f0a9bc491657488e4a-Abstract-Datasets_and_Benchmarks_Track.html}
}

@inproceedings{liu2024iagents,
  title={Autonomous Agents for Collaborative Task under Information Asymmetry},
  author={Liu, Wei and Wang, Chenxi and Wang, Yifei and Xie, Zihao and Qiu, Rennai and Dang, Yufan and Du, Zhuoyun and Chen, Weize and Yang, Cheng and Qian, Chen},
  booktitle={Advances in Neural Information Processing Systems},
  volume={37},
  year={2024},
  doi={10.52202/079017-0090},
  url={https://papers.nips.cc/paper_files/paper/2024/hash/0534abc9e6db91683d82186ef0d68202-Abstract-Conference.html}
}

@inproceedings{zhu2025multiagentbench,
  title={{MultiAgentBench}: Evaluating the Collaboration and Competition of {LLM} Agents},
  author={Zhu, Kunlun and Du, Hongyi and Hong, Zhaochen and Yang, Xiaocheng and Guo, Shuyi and Wang, Zhe and Wang, Zhenhailong and Qian, Cheng and Tang, Xiangru and Ji, Heng and You, Jiaxuan},
  booktitle={Proceedings of the 63rd Annual Meeting of the Association for Computational Linguistics (Volume 1: Long Papers)},
  pages={8580--8622},
  year={2025},
  doi={10.18653/v1/2025.acl-long.421},
  url={https://aclanthology.org/2025.acl-long.421/}
}

@inproceedings{kim2023fantom,
  title={{FANToM}: A Benchmark for Stress-testing Machine Theory of Mind in Interactions},
  author={Kim, Hyunwoo and Sclar, Melanie and Zhou, Xuhui and Bras, Ronan and Kim, Gunhee and Choi, Yejin and Sap, Maarten},
  booktitle={Proceedings of the 2023 Conference on Empirical Methods in Natural Language Processing},
  year={2023},
  url={https://aclanthology.org/2023.emnlp-main.890/}
}

@inproceedings{li2026hiddenbench,
  title     = {Systematic Failures in Collective Reasoning under Distributed Information in Multi-Agent {LLM}s},
  author    = {Li, Yuxuan and Naito, Aoi and Shirado, Hirokazu},
  booktitle = {Proceedings of the 43rd International Conference on Machine Learning},
  year      = {2026},
  url       = {https://arxiv.org/abs/2505.11556}
}

@misc{openai2026gpt56sol,
  author={{OpenAI}},
  title={{GPT-5.6 Sol Model}},
  year={2026},
  url={https://developers.openai.com/api/docs/models/gpt-5.6-sol}
}

@misc{anthropic2026claudeopus5,
  author={{Anthropic}},
  title={Introducing {Claude Opus 5}},
  year={2026},
  month=jul,
  url={https://www.anthropic.com/news/claude-opus-5}
}

@misc{deepseek2026v41flash,
  author={{DeepSeek-AI}},
  title={{DeepSeek-V4.1-Flash Release}},
  year={2026},
  month=sep,
  url={https://api-docs.deepseek.com/updates/}
}

@article{glm5team2026glm5,
  author={{GLM-5 Team}},
  title={{GLM-5: From Vibe Coding to Agentic Engineering}},
  journal={arXiv preprint arXiv:2602.15763},
  year={2026},
  url={https://arxiv.org/abs/2602.15763}
}

@misc{meta2026museglimmer,
  author={{Meta Superintelligence Lab}},
  title={{Muse-Glimmer-30B} Model Card},
  year={2026},
  month=aug,
  url={https://huggingface.co/meta-models/Muse-Glimmer-30B}
}

@article{gemmateam2026gemma4,
  author={{Gemma Team}},
  title={{Gemma 4 Technical Report}},
  journal={arXiv preprint arXiv:2607.02770},
  year={2026},
  url={https://arxiv.org/abs/2607.02770}
}

@misc{google2026gemini38flash,
  author={{Google}},
  title={{Gemini 3.8 Flash}},
  year={2026},
  url={https://ai.google.dev/gemini-api/docs/models/gemini-3.8-flash}
}

@misc{tencent2026hy4,
  author={{Tencent Hy Team}},
  title={{Hy4} Model Card},
  year={2026},
  month=nov,
  url={https://huggingface.co/tencent/Hy4}
}

@misc{tencent2026hy4flash,
  author={{Tencent Hy Team}},
  title={{Hy4-Flash} Model Card},
  year={2026},
  month=nov,
  url={https://huggingface.co/tencent/Hy4-flash}
}

@misc{qwen2026qwen4max,
  author={{Qwen Team}},
  title={{Qwen4-Max}: Your New Teammate Has Arrived},
  year={2026},
  month=nov,
  url={https://qwen.ai/blog?id=qwen4-max}
}

@misc{qwen2026qwen38,
  author={{Qwen Team}},
  title={{Qwen3.8-27B} Model Card},
  year={2026},
  url={https://huggingface.co/Qwen/Qwen3.8-27B}
}

@misc{qwen2026qwen35,
  author={{Qwen Team}},
  title={{Qwen3.5-9B} Model Card},
  year={2026},
  url={https://huggingface.co/Qwen/Qwen3.5-9B}
}

@misc{mistralai2026mistralsmall4,
  author={{Mistral AI}},
  title={{Mistral Small 4 119B A6B} Model Card},
  year={2026},
  url={https://huggingface.co/mistralai/Mistral-Small-4-119B-2603}
}

@inproceedings{kwon2023vllm,
  title={Efficient Memory Management for Large Language Model Serving with {PagedAttention}},
  author={Kwon, Woosuk and Li, Zhuohan and Zhuang, Siyuan and Sheng, Ying and Zheng, Lianmin and Yu, Cody Hao and Gonzalez, Joseph E. and Zhang, Hao and Stoica, Ion},
  booktitle={Proceedings of the 29th Symposium on Operating Systems Principles},
  year={2023},
  doi={10.1145/3600006.3613165},
  url={https://arxiv.org/abs/2309.06180}
}

@misc{deepseek2026v4flash0731,
  author={{DeepSeek-AI}},
  title={{DeepSeek-V4-Flash Update}},
  year={2026},
  month=jul,
  url={https://api-docs.deepseek.com/updates/}
}

@inproceedings{li2025agent,
  title={Agent-oriented planning in multi-agent systems},
  author={Li, Ao and Xie, Yuexiang and Li, Songze and Tsung, Fugee and Ding, Bolin and Li, Yaliang},
  booktitle={International Conference on Learning Representations},
  volume={2025},
  pages={19495--19517},
  year={2025}
}

@inproceedings{yun2026graph,
  title={Graph-of-agents: A graph-based framework for multi-agent LLM collaboration},
  author={Yun, Sukwon and Peng, Jie and Li, Pingzhi and Fan, Wendong and Chen, Jie and Zou, James Y and Li, Guohao and Chen, Tianlong},
  booktitle={International Conference on Learning Representations},
  volume={2026},
  pages={19745--19760},
  year={2026}
}

@inproceedings{chen2025internet,
  title={Internet of agents: Weaving a web of heterogeneous agents for collaborative intelligence},
  author={Chen, Weize and You, Ziming and Li, Ran and Qian, Chen and Zhao, Chenyang and Yang, Cheng and Xie, Ruobing and Liu, Zhiyuan and Sun, Maosong and others},
  booktitle={International Conference on Learning Representations},
  volume={2025},
  pages={36374--36411},
  year={2025}
}

@article{lingard2004communication,
  title={Communication failures in the operating room: an observational classification of recurrent types and effects},
  author={Lingard, Lorelei and Espin, Sherry and Whyte, Sarah and Regehr, Glenn and Baker, G Ross and Reznick, Richard and Bohnen, John and Orser, Beverly and Doran, Diane and Grober, Ellen},
  journal={Quality and safety in Health care},
  volume={13},
  number={5},
  pages={330--334},
  year={2004},
  publisher={BMJ Publishing Group Ltd}
}

@article{reason2000human,
  title={Human error: models and management},
  author={Reason, James},
  journal={Bmj},
  volume={320},
  number={7237},
  pages={768--770},
  year={2000},
  publisher={British Medical Journal Publishing Group}
}

@inproceedings{liu2025mcpeval,
  title={Mcpeval: Automatic mcp-based deep evaluation for ai agent models},
  author={Liu, Zhiwei and Qiu, Jielin and Wang, Shiyu and Zhang, Jianguo and Liu, Zuxin and Ram, Roshan and Chen, Haolin and Yao, Weiran and Heinecke, Shelby and Savarese, Silvio and others},
  booktitle={Proceedings of the 2025 Conference on Empirical Methods in Natural Language Processing: System Demonstrations},
  pages={373--402},
  year={2025}
}

@article{wang2025odysseybench,
  title={OdysseyBench: Evaluating LLM Agents on Long-Horizon Complex Office Application Workflows},
  author={Wang, Weixuan and Han, Dongge and Diaz, Daniel Madrigal and Xu, Jin and R{\"u}hle, Victor and Rajmohan, Saravan},
  journal={arXiv preprint arXiv:2508.09124},
  year={2025}
}

@inproceedings{PARTNR,
  author = {Matthew Chang and Gunjan Chhablani and Alexander Clegg and Mikael Dallaire Cote and Ruta Desai and Michal Hlavac and Vladimir Karashchuk and Jacob Krantz and Roozbeh Mottaghi and Priyam Parashar and Siddharth Patki and Ishita Prasad and Xavier Puig and Akshara Rai and Ram Ramrakhya and Daniel Tran and Joanne Truong and John M. Turner and Eric Undersander and Tsung-Yen Yang},
  title = {PARTNR: A Benchmark for Planning and Reasoning in Embodied Multi-agent Tasks},
  booktitle = {International Conference on Learning Representations (ICLR)},
  note = {alphabetical author order},
  year = {2025}
}

@inproceedings{huang2025resilience,
  title         = {On the Resilience of {LLM}-Based Multi-Agent Collaboration with Faulty Agents},
  author        = {Huang, Jen-Tse and Zhou, Jiaxu and Jin, Tailin and Zhou, Xuhui and Chen, Zixi and Wang, Wenxuan and Yuan, Youliang and Lyu, Michael and Sap, Maarten},
  booktitle     = {Proceedings of the 42nd International Conference on Machine Learning},
  series        = {Proceedings of Machine Learning Research},
  volume        = {267},
  pages         = {26202--26226},
  publisher     = {PMLR},
  year          = {2025},
  url           = {https://proceedings.mlr.press/v267/huang25ay.html}
}

@misc{jia2026masfire,
  title         = {{MAS-FIRE}: Fault Injection and Reliability Evaluation for {LLM}-Based Multi-Agent Systems},
  author        = {Jia, Jin and Deng, Zhiling and Chen, Zhuangbin and Wang, Yingqi and Zheng, Zibin},
  year          = {2026},
  howpublished  = {arXiv preprint arXiv:2602.19843},
  eprint        = {2602.19843},
  archivePrefix = {arXiv},
  primaryClass  = {cs.SE},
  doi           = {10.48550/arXiv.2602.19843},
  url           = {https://arxiv.org/abs/2602.19843}
}

@inproceedings{zhang2026agentracer,
  title         = {{AgenTracer}: Who Is Inducing Failure in the {LLM} Agentic Systems?},
  author        = {Zhang, Guibin and Wang, Junhao and Chen, Junjie and Zhou, Wangchunshu and Wang, Kun and Yan, Shuicheng},
  booktitle     = {The Fourteenth International Conference on Learning Representations},
  year          = {2026},
  url           = {https://openreview.net/forum?id=l05DseqvuD}
}

@inproceedings{sun2024ToolsFail,
  title         = {Tools Fail: Detecting Silent Errors in Faulty Tools},
  author        = {Sun, Jimin and Min, So Yeon and Chang, Yingshan and Bisk, Yonatan},
  booktitle     = {Proceedings of the 2024 Conference on Empirical Methods in Natural Language Processing},
  pages         = {14272--14289},
  publisher     = {Association for Computational Linguistics},
  year          = {2024},
  doi           = {10.18653/v1/2024.emnlp-main.790},
  url           = {https://aclanthology.org/2024.emnlp-main.790/}
}

@inproceedings{wang2025HellOrHighWater,
  title         = {Hell or High Water: Evaluating Agentic Recovery from External Failures},
  author        = {Wang, Andrew and Hager, Sophia and Asija, Adi and Khashabi, Daniel and Andrews, Nicholas},
  booktitle     = {Second Conference on Language Modeling},
  year          = {2025},
  url           = {https://openreview.net/forum?id=Zk224WPT42}
}

@inproceedings{zhao2026AskBench,
  title         = {When and What to Ask: {AskBench} and Rubric-Guided {RLVR} for {LLM} Clarification},
  author        = {Zhao, Jiale and Fang, Ke and Cheng, Lu},
  booktitle     = {Findings of the Association for Computational Linguistics: ACL 2026},
  pages         = {17120--17140},
  publisher     = {Association for Computational Linguistics},
  year          = {2026},
  doi           = {10.18653/v1/2026.findings-acl.845},
  url           = {https://aclanthology.org/2026.findings-acl.845/}
}

@misc{wang2026AgentNoiseBench,
  title         = {{AgentNoiseBench}: Benchmarking Robustness of Tool-Using {LLM} Agents Under Noisy Condition},
  author        = {Wang, Ruipeng and Chen, Yuxin and Wang, Yukai and Wu, Chang and Fang, Junfeng and Cai, Xiaodong and Gu, Qi and Su, Hui and Zhang, An and Wang, Xiang and Cai, Xunliang and Chua, Tat-Seng},
  year          = {2026},
  howpublished  = {arXiv preprint arXiv:2602.11348},
  eprint        = {2602.11348},
  archivePrefix = {arXiv},
  primaryClass  = {cs.AI},
  url           = {https://arxiv.org/abs/2602.11348}
}

@inproceedings{chen2024RGB,
  title         = {Benchmarking Large Language Models in Retrieval-Augmented Generation},
  author        = {Chen, Jiawei and Lin, Hongyu and Han, Xianpei and Sun, Le},
  booktitle     = {Proceedings of the AAAI Conference on Artificial Intelligence},
  volume        = {38},
  pages         = {17754--17762},
  year          = {2024},
  doi           = {10.1609/aaai.v38i16.29728},
  url           = {https://ojs.aaai.org/index.php/AAAI/article/view/29728}
}

@article{zhang2025agent,
  title={Which agent causes task failures and when? on automated failure attribution of llm multi-agent systems},
  author={Zhang, Shaokun and Yin, Ming and Zhang, Jieyu and Liu, Jiale and Han, Zhiguang and Zhang, Jingyang and Li, Beibin and Wang, Chi and Wang, Huazheng and Chen, Yiran and others},
  journal={arXiv preprint arXiv:2505.00212},
  year={2025}
}
